\documentclass[letterpaper]{article}

\usepackage{aaai2027}
\usepackage{times}
\usepackage{helvet}
\usepackage{courier}
\usepackage[hyphens]{url}
\usepackage{graphicx}
\usepackage{natbib}
\usepackage{caption}
\nocopyright

\usepackage{booktabs}
\usepackage{tabularx}
\usepackage{array}
\usepackage{multirow}
\usepackage{algorithm}
\usepackage{algorithmic}
\usepackage{booktabs}
\usepackage{multirow}
\usepackage{makecell}
\usepackage{pifont}
\usepackage{amsmath}
\usepackage{placeins}

\usepackage{newfloat}
\usepackage{listings}
\DeclareCaptionStyle{ruled}{labelfont=normalfont,labelsep=colon,strut=off}
\floatstyle{ruled}
\newfloat{listing}{tb}{lst}{}
\floatname{listing}{Listing}

\newcommand{\yes}{\ding{51}}
\newcommand{\no}{\ding{55}}

\title{\textit{Thinking Once} Is Enough: Intermediate-Layer Evidence Routing \\for High-Resolution VQA}

\author{
    Zhongkuan Mao\equalcontrib\textsuperscript{\rm 1},
    Xianjie Liu\equalcontrib\textsuperscript{\rm 2},
    Tianyu Meng\textsuperscript{\rm 3},
    Yidong Wang\textsuperscript{\rm 4},
    Wenzhuo Zhao\textsuperscript{\rm 2},
    Ronghao Xian\textsuperscript{\rm 2},
    Yao Jiang\textsuperscript{\rm 2},
    Fei Shen\textsuperscript{\rm 5},
    Junfeng Fang\textsuperscript{\rm 5},
    Yong Dai\textsuperscript{\rm 6},
    Yi Zhang\textsuperscript{\rm 6},
    Keren Fu\textsuperscript{\rm 1,\rm 2}\thanks{Corresponding author.}
}

\affiliations{
    \textsuperscript{\rm 1}National Key Laboratory of Fundamental Science on Synthetic Vision, Sichuan University\\
    \textsuperscript{\rm 2}College of Computer Science, Sichuan University\\
    \textsuperscript{\rm 3}Beihang University
    \textsuperscript{\rm 4}Peking University
    \textsuperscript{\rm 5}National University of Singapore
    \textsuperscript{\rm 6}X-Humanoid \\
    mttjelly0628@163.com
}

\begin{document}

\maketitle

\begin{abstract}

High-resolution visual question answering (HR-VQA) is often treated as a problem of insufficient evidence acquisition, where failing multimodal large language models must inspect images again through cropping, re-encoding, or multi-round search. We show that this view is incomplete: in many cases, fine-grained evidence has already survived visual encoding and become identifiable and influential within an intermediate-layer routing window, but is later diluted before answer generation. We propose Thinking-Once, a \textbf{training-free, single-visual-pass} evidence-routing method that reconstructs question-conditioned attention at this window, preserves core entity tokens and compact background context, and routes this evidence to later layers without extra visual encoding. Across five base models, Thinking-Once consistently improves or matches the corresponding base setting, increasing the average scores on V$^*$Bench, HRBench-4K, and HRBench-8K by \textit{+3.1}, \textit{+3.0}, and \textit{+2.7} points while reducing the average peak memory by about 4,GB. On Qwen2.5-VL-7B, it improves the three benchmarks by \textit{+9.9}, \textit{+4.6}, and \textit{+5.5} points, raising the cross-benchmark mean from 72.5 to 79.1. With the ZwZ-8B base model, Thinking-Once reaches a mean score of 82.7. Against 11 open-source HR-VQA baselines, it obtains the best or tied-best score on all three benchmark averages and the best overall mean; for example, compared with DeepScan, it reduces V$^*$Bench inference time by \textbf{97.2\%} while improving the cross-benchmark mean from 77.8 to 79.1. These results show that HR-VQA can be improved by routing already encoded evidence rather than repeatedly acquiring new visual inputs. Code is available in the appendix.

% The code will be public available.

\end{abstract}

\section{1. Introduction}

Multimodal large language models (MLLMs) have achieved strong performance on standard visual understanding benchmarks~\cite{alayrac2022flamingo,li2023blip-2,liu2023Llava,dai2023instructblip,zhu2024minigpt,wang2024exploring,liu2025nexus,qwen2.5-vl,qwen3-vl,bai2023qwenvl}, yet they continue to struggle with high-resolution and fine-grained tasks where the model must recognize small objects, read localized text, or deduce spatial relationships. The core difficulty lies in the need to precisely localize subtle, question-relevant visual evidence amid overwhelming background interference and then leverage these localized cues for subsequent perception and reasoning. When high-resolution images are encoded directly, the resulting large volume of visual tokens often dilutes attention to question-relevant evidence, causing fine-grained details to be overshadowed by redundant global context.

\begin{figure}[t]
    \centering
    \includegraphics[width=\linewidth]{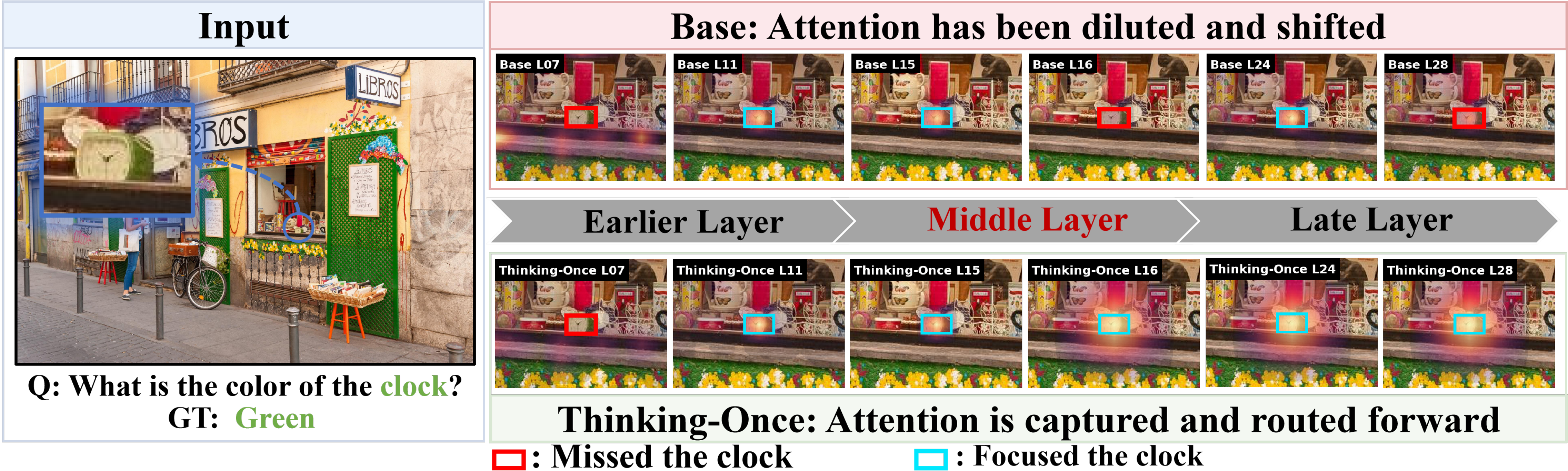}
    \caption{
    The base model gradually loses focus on the question-relevant region, whereas Thinking-Once more effectively captures the relevant evidence at an intermediate layer and preserves it throughout subsequent reasoning.
    }
    \label{fig:intro_comparison}
\end{figure}

Existing strategies follow two directions. \emph{External evidence reacquisition} methods obtain extra visual evidence through multi-turn visual search, crop/zoom-in, region re-encoding, or training-time zooming distillation~\cite{ToMe2023,khayatkhoei2025mllms,zhang2025chain,shen2025zoomeye,zheng2025deepeyes,liu2025hide,li2026reliable,wei2026zooming,jiang2026vlm,DeepScan2026}. They can recover missing details, but they do so by changing the visual input pipeline, which leads to substantial inference cost. In contrast, \emph{compression-first token reduction} methods improve efficiency by pruning or merging visual tokens~\cite{chen2024image,yang2025visionzip,shang2025llava,li2025tokenpacker,V2Drop2026,liu2026hiprune,han2026filter,zhang2026vscan}. Their generic redundancy criteria can often remove entity or contextual tokens that remain necessary for relation-sensitive HR-VQA. These limitations motivate a different perspective: rather than relying solely on additional observation or token reduction, we more effectively route the critical question-relevant evidence that has already emerged in intermediate layers within a single visual pass.

Our observations in the \emph{Observations and Analyses} section collectively support a different view. As shown in Fig.~\ref{fig:intro_comparison}, MLLMs can already activate question-relevant regions at intermediate layers, but this evidence may become diluted or displaced during subsequent reasoning. Meanwhile, visual embeddings produced by the vision encoder can retain fine-grained information from the original image~\cite{meituanlongcatteam2026longcatnextlexicalizingmodalitiesdiscrete}. These observations suggest that HR-VQA failures are not always caused by missing visual evidence; they can potentially also arise from the ineffective preservation and transmission of evidence that has been encoded and localized. We therefore shift the focus from repeatedly acquiring new visual inputs or indiscriminately compressing tokens to routing question-relevant evidence through later layers, leading to the \emph{Thinking-Once evidence-routing} paradigm.

Thinking-Once emphasizes evidence filtering and delivery within a single visual processing flow, relying on one full-image visual encoding and one multimodal forward pass. We extract visual tokens from an intermediate layer and utilize the cross-modal attention distributions that develop at this stage to isolate core entity tokens highly aligned with the current question, while retaining contextual background tokens that support relation inference, attribute discrimination, and scene parsing, and route the filtered evidence set toward deeper network layers. Through this design, we reformulate visual information selection in HR-VQA as \emph{query-conditioned reasoning-evidence routing}, which explicitly preserves the integrity of visual evidence required for downstream reasoning within a single-visual-pass framework.

The main contributions of this paper are as follows:
\begin{itemize}
    \item \textbf{Insightful Observations:} We identify the ineffective transmission of question-relevant visual evidence that has already emerged in intermediate layers as an important and underexplored source of HR-VQA failure, complementing the conventional evidence-acquisition view.
    \item \textbf{Streamlined Paradigm:} We propose the Thinking-Once evidence routing mechanism, which enables high-resolution reasoning with one full-image encoding and one forward pass by routing intermediate-layer visual evidence without constructing extra visual inputs.
    \item \textbf{Systematic Method and Validation:} We design a query-conditioned evidence routing mechanism to transmit core and background tokens. Experiments and visualizations demonstrate that our approach significantly outperforms representative methods from existing paradigms within a simpler single-visual-pass reasoning process.
\end{itemize}

\section{2. Related Work}
\subsection{2.1. High-Resolution VQA and Existing Paradigms}

High-resolution visual question answering (HR-VQA) evaluates whether MLLMs can perceive and reason over fine-grained evidence in high-resolution scenes, including small objects, localized text, detailed attributes, and spatial relations. Benchmarks such as V$^*$Bench~\cite{vstar2024} and HRBench~\cite{hrbench2025} show that existing MLLMs still struggle to capture question-relevant local evidence when high-resolution inputs produce long visual token sequences. Existing HR-VQA methods mainly enhance local perception through additional visual operations. \emph{Visual-search} methods locate relevant regions with tool use or multi-step observation~\cite{zhang2025chain,li2025dyfo};
\emph{crop/zoom-in} methods crop or re-encode selected sub-images for local details~\cite{khayatkhoei2025mllms,shen2025zoomeye,zheng2025deepeyes, liu2025hide,li2026reliable,jiang2026vlm,DeepScan2026}; and \emph{ZwZ} internalizes the ability to zoom in through training-time distillation~\cite{wei2026zooming}. Although effective, these methods require extra computation, additional visual inputs, or complex training, and overlook evidence already formed in intermediate layers. In contrast, we more directly route such internal evidence with one full-image visual encoding and one multimodal forward pass.

\subsection{2.2. Token Reduction and Evidence Routing}

Efficiency-oriented methods reduce the cost of high-resolution MLLMs by pruning or merging visual tokens according to attention scores, token similarity, or saliency~\cite{ToMe2023,liu2026hiprune,yang2025visionzip,chen2024image,shang2025llava,li2025tokenpacker,han2026filter,zhang2026vscan}. Although effective, these methods primarily identify redundant tokens for removal, and their generic selection criteria may discard background or contextual tokens needed for attribute recognition, spatial reasoning, and relation understanding. Thinking-Once is related to this line of work at the implementation level because it also shortens later-layer sequences, but differs in objective and selection principle: instead of optimizing token reduction itself, it constructs a question-conditioned evidence set through independent minimum-coverage selection for each routing query and more explicitly preserves structured contextual support with background summaries. Thus, compression emerges in practice as a consequence of evidence routing, rather than being the sole design objective.

\section{3. Observations and Analysis}
\label{sec:analysis}

HR-VQA is treated as an evidence-acquisition problem: when a model fails, it should see the image again at a finer scale. This explanation is incomplete if the required evidence has already entered the model but is later diluted, misallocated, or discarded. So we diagnose HR-VQA with three questions: \emph{Is the evidence available after visual encoding? Where and when does it become useful? What information must remain around the evidence core?} The answers distinguish evidence availability from evidence utilization and provide direct design constraints for Thinking-Once.

\subsection{3.1. Encoded Visual Representations Retain Fine-Grained Evidence}

Single-visual-pass routing is viable if the initial encoding retains information needed for later reasoning. LongCat-Next~\cite{meituanlongcatteam2026longcatnextlexicalizingmodalitiesdiscrete} shows that lightweight decoders can reconstruct image structure, object boundaries, and spatial layout from frozen vision-encoder outputs, which indicates that visual embeddings retain recoverable fine-grained structure. Reconstructability establishes a narrower but critical condition, namely that local information can survive visual encoding and remain available to subsequent layers. Complementary evidence from interleaved visual reasoning, vision tool-use training, and recent analyses of RL-induced reasoning changes~\cite{yang2026position,ma2026does} points to the same, and we summarize it in Appendix B.

These observations identify an underexplored failure mode, consistent with recent evidence that large models may internally encode relevant evidence even when it is not surfaced in the final prediction~\cite{liu2025selfelicit}. In a high-resolution image, a few relevant tokens compete with a larger set of unrelated tokens throughout language-side reasoning, so that an encoded target can be diluted before answer generation. The resulting question is therefore whether the model can allocate sufficient downstream computation to evidence that is already available, in addition to whether it can acquire that evidence. Thinking-Once targets this utilization-limited regime, and it does not claim to replace visual tools when the original encoding genuinely lacks the required information.

\subsection{3.2. Intermediate Layers Define an Evidence-Routing Window}

\begin{figure}[t]
    \centering
    \includegraphics[width=\linewidth]{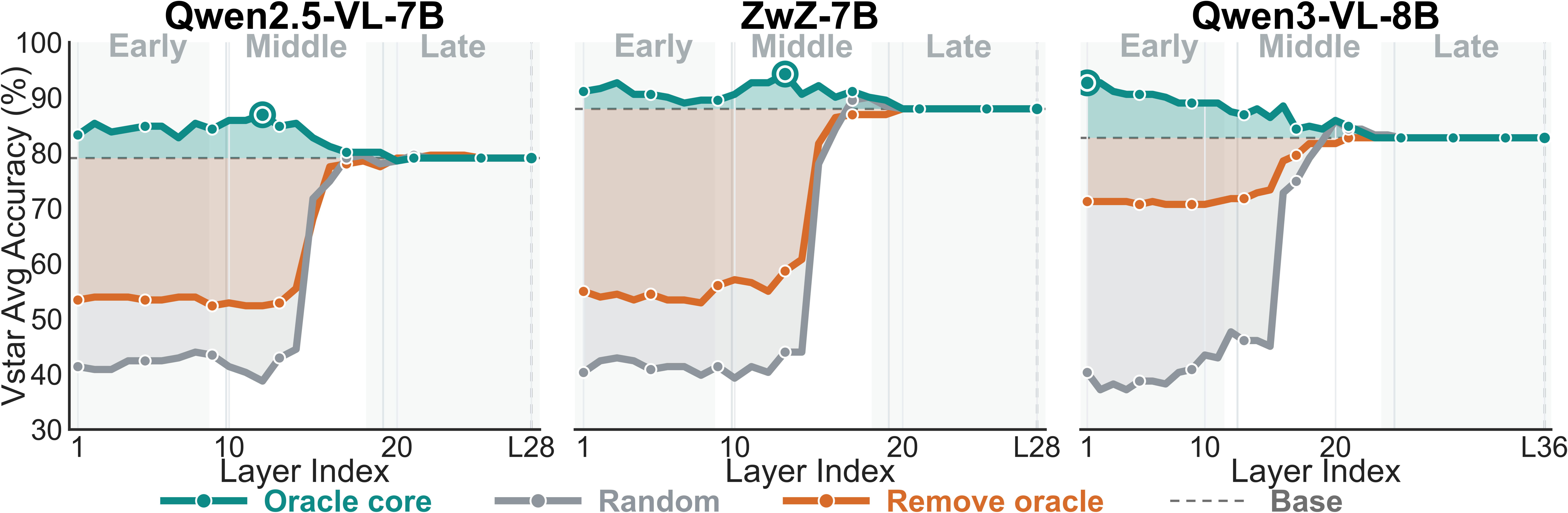}
    \caption{
    Layer-wise oracle token intervention on V$^*$Bench. Oracle core tokens consistently outperform random tokens across most layers, while removing them degrades performance, especially in early and intermediate layers.
}
    \label{fig:oracle_intervention}
\end{figure}

We localize the evidence within the visual sequence using three layer-wise oracle interventions. \emph{Random Tokens} retain a random set with the same budget as the ground-truth-box tokens; \emph{Oracle Core Tokens} retain only tokens inside the box; and \emph{Remove Oracle Tokens} retain all tokens except those inside the box. Applying these interventions at transformer layers separates factors otherwise mixed: token budget, token identity, and intervention depth. This design allows us to ask whether the identity of the retained evidence influences the answer, which goes beyond asking whether fewer tokens can work. Figure~\ref{fig:oracle_intervention} reveals an asymmetric evidence distribution. Under the same budget, Oracle Core Tokens outperform Random Tokens and can even exceed the base model, whereas removing the oracle tokens cannot be compensated for by retaining the much larger available background set. Entity tokens are therefore consistently information-dense and hard to replace: HR-VQA depends substantially more on preserving the right tokens than simply preserving many tokens.

The two oracle comparisons provide complementary evidence. The advantage over Random Tokens supports \textbf{\emph{sufficiency}}, since a compact entity-centered subset can support strong reasoning when it contains the relevant evidence. The failure of Remove Oracle Tokens supports \textbf{\emph{necessity}}, since abundant non-target information cannot reliably reconstruct evidence removed from the entity region. Because Random Tokens use the same budget as Oracle Core Tokens and Remove Oracle Tokens keep more tokens overall, the result cannot be explained by token count alone. What matters is the identity of the retained tokens and the precise layer at which their influence is most effectively redirected.

The intervention effect is also layer-dependent, which lets us locate where routing should happen. Its advantage is strongest when question--entity correspondence has emerged while it can still influence the remaining computation, and interventions at deeper layers converge toward the base model as evidence becomes absorbed into later hidden states. This more clearly defines an \emph{evidence-routing window}. At early layers, visual detail is available while question-conditioned correspondence may still be weak. At late layers, the model has formed richer multimodal states, while too little computation remains for an intervention to redirect reasoning. Intermediate layers provide the useful overlap, where evidence has become identifiable and sufficiently influential under intervention, which is the operating regime required by single-visual-pass routing. The earliest oracle optimum should not be read as the practical routing layer, because ground-truth boxes provide perfect localization that is unavailable at inference time. A usable routing layer must jointly provide reliable question-conditioned localization and sufficient remaining depth for the routed evidence to affect the answer.

\subsection{3.3. Core Evidence Requires Contextual Support}
\begin{figure}[t]
    \centering
    \includegraphics[width=\linewidth]{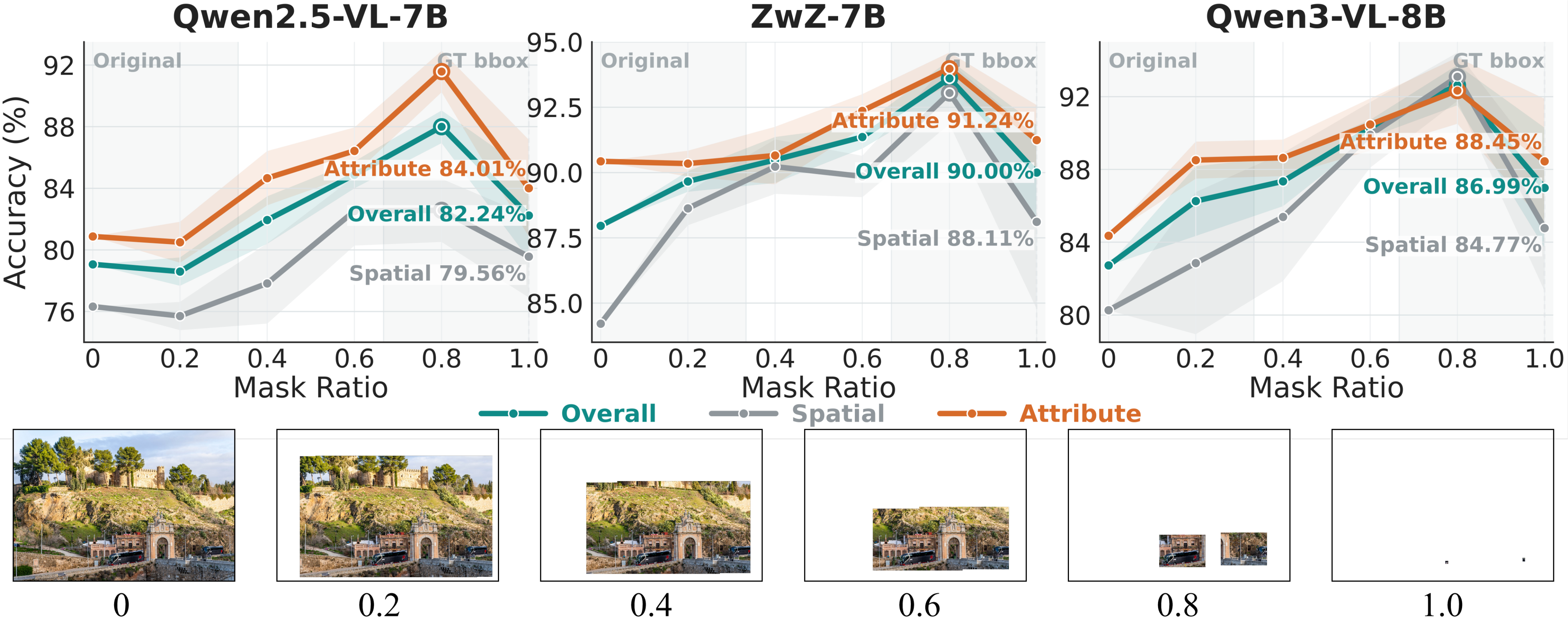}
    \caption{Effect of progressive background masking on HR-VQA performance. We gradually vary the background mask ratio from the original full image to the GT-bbox-only input.}
    \label{fig:bg_masking}
\end{figure}

Entity tokens form the evidence core, but retaining this core is insufficient. Figure~\ref{fig:bg_masking} shows a non-monotonic effect of background removal: moderate masking improves accuracy, whereas aggressive masking degrades it. This indicates that background tokens contain both interference and supporting context, including spatial references, relational cues, and attribute-comparison evidence. The optimal representation is selective rather than minimal. This also explains why object-only cropping can be brittle: many HR-VQA questions refer to an entity but ask about relations, relative positions, or comparisons whose evidence lies outside the entity box. Removing all surrounding tokens preserves the referent but destroys part of the structure needed to interpret it.

\begin{figure}[t]
    \centering
    \includegraphics[width=\linewidth]{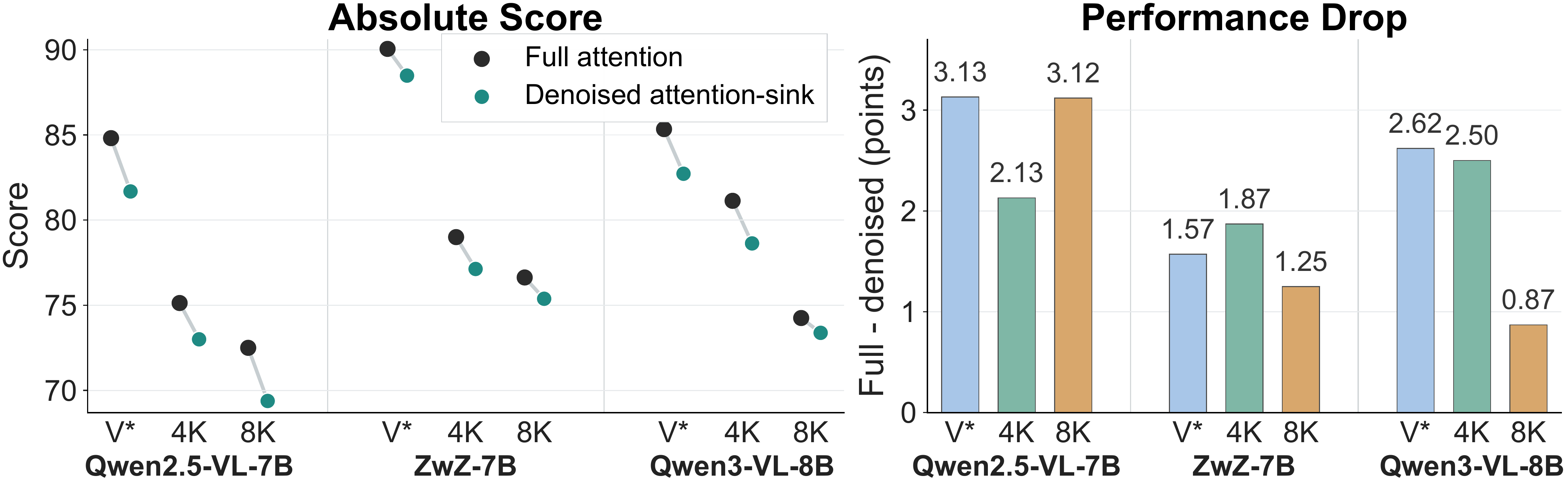}
    \caption{Effect of attention-sink suppression on HR-VQA performance. The consistent degradation across models and benchmarks suggests that these tokens contribute useful contextual or aggregation signals during downstream reasoning.}
    \label{fig:attn_sink}
\end{figure}

The attention-sink intervention in Figure~\ref{fig:attn_sink} further shows that localization and utility are not equivalent. Although such tokens may distort attention-based cropping~\cite{liu2025hide}, suppressing them reduces accuracy across MLLMs and benchmarks. This suggests that spatially non-discriminative tokens can still carry contextual, aggregation, or information-transfer signals, consistent with prior findings on attention-sink and register-like tokens in
Transformers~\cite{xiao2024efficient,darcet2024vision,kang2025see}. Token utility should be judged by its contribution to later reasoning, not only by whether it maps onto a visible object.

The masking and sink interventions show that evidence in HR-VQA is structured rather than sparse. Spatially adjacent background provides explicit context for relations and comparisons, while attention-sink states may provide implicit context through aggregation and information transfer. Keeping the entire visual sequence preserves context but leaves the evidence core exposed to high-resolution redundancy; keeping only the most salient entity removes interference but loses the support needed to interpret that entity. Useful routing lies between these extremes: it should effectively preserve the smallest evidence structure that supports the question, rather than maximizing retention or compression.

\subsection{3.4. From Analysis to Design Principles}

The evidence-allocation view clarifies the applicability boundary of Thinking-Once. Routing cannot recover information that the vision encoder has irreversibly removed, nor can it solve cases where the relevant region is never represented in the initial encoding. It is therefore complementary to cropping, zooming, or visual search, which remain useful when new visual observations are needed. Thinking-Once instead targets the utilization-limited regime, where the evidence is encoded but not effectively preserved or allocated. We provide a more detailed discussion of these boundaries and their testable predictions in Appendices B and F.

\begin{figure}[t]
    \centering
    \includegraphics[width=\columnwidth]{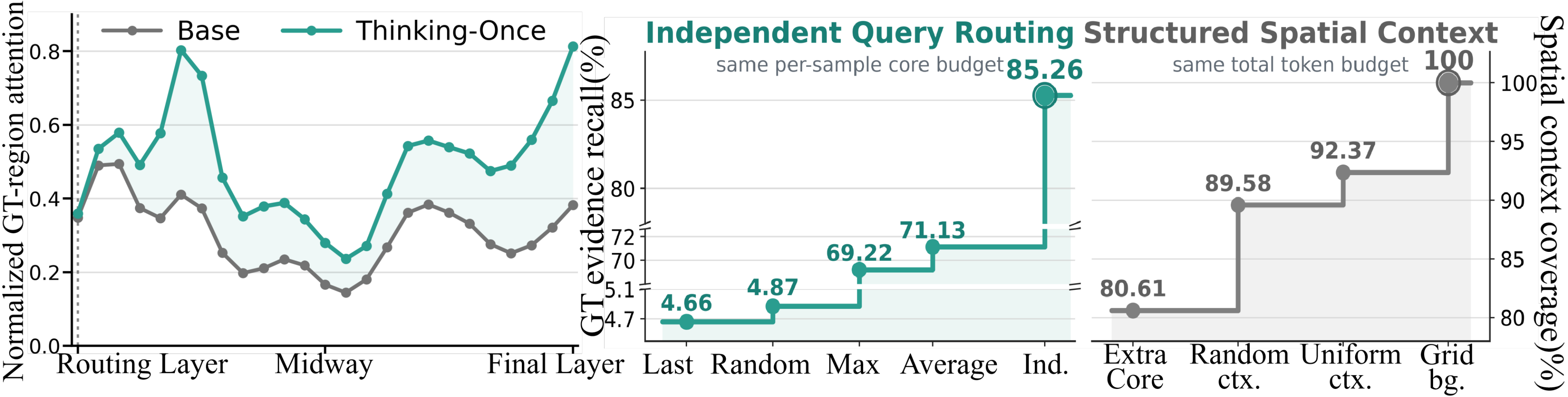}
    \caption{
    Left: GT-evidence recall across query-routing strategies under the same per-sample core-token budget.
    Right: spatial-context coverage across variants under the same total-token budget.
    Ind. denotes Independent.
    }
    \label{fig:routing_design_validation}
\end{figure}

The above analyses are translated into four design requirements. Routing should operate on internal representations, be conditioned on the question, occur at intermediate layers where evidence is identifiable and actionable, and preserve compact contextual support rather than aggressively pruning every non-entity token. Figure~\ref{fig:routing_design_validation} further validates the two key design choices of Thinking-Once under matched token budgets. Under the same per-sample core-token budget, independent query routing achieves the highest GT-evidence recall, indicating that preserving evidence separately for different routing queries avoids the information suppression caused by query aggregation. Under the same total-token budget, structured grid background summaries provide the highest spatial-context coverage, showing that the spatial organization of contextual tokens is more effective than simply adding core tokens or allocating context randomly or uniformly. These results support independent evidence selection and structured background preservation as complementary components of the proposed routing mechanism.

Together, these requirements define Thinking-Once: a single-visual-pass mechanism that preserves the structure of already encoded, question-aligned evidence and routes later-layer computation more effectively toward it.

\begin{figure*}[t]
    \centering
    \includegraphics[width=\textwidth]{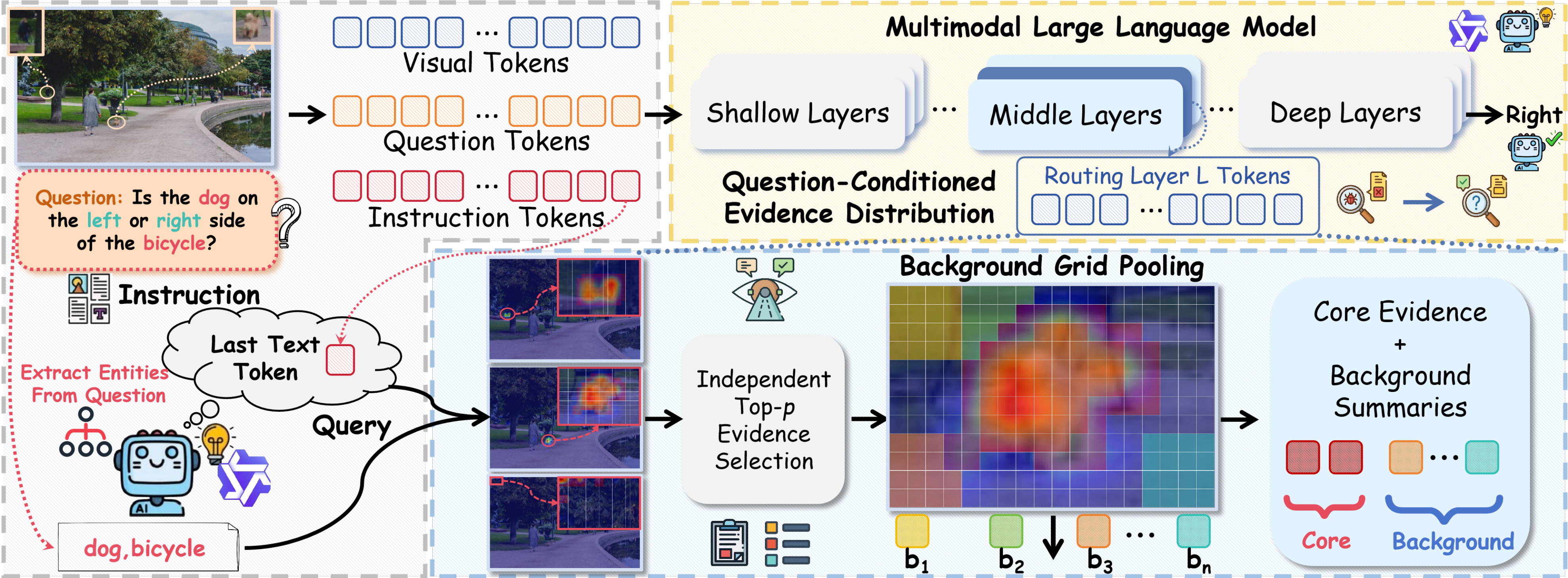}
    \caption{
    Overview of Thinking-Once. At a routing layer, question-conditioned queries independently select core evidence, while background grid pooling summarizes visual context before routing both to deeper layers within a single visual pass.
    }
    \label{fig:method_overview}
\end{figure*}

\section{4. Methodology}
\label{sec:method}

As illustrated in Figure~\ref{fig:method_overview}, Thinking-Once performs question-guided evidence routing at inference time with one full-image visual encoding and one multimodal forward pass. More specifically, given a high-resolution image and a question, an MLLM with $D$ transformer layers first runs the full multimodal sequence to an intermediate routing layer $L<D$:
\begin{equation}
    H^{L}=\mathrm{SeqMerge}(V^{L};T^{L}), \qquad
    V^{L}=\{v^{L}_{i}\}_{i=1}^{N},
    \label{eq:method_state}
\end{equation}
where $V^{L}$ and $T^{L}$ denote the visual and text hidden states, and $\mathrm{SeqMerge}$ preserves their original prompt order. At layer $L$, Thinking-Once replaces the visual subsequence with a compact evidence representation $E^{L}$ while keeping the text sequence unchanged. The image is encoded only once, with no crop, sub-image re-encoding, or additional visual observation. Derivations are provided in Appendix A.

\subsection{4.1. Question-Conditioned Evidence Distribution}
\label{sec:evidence_scoring}

Different question parts may require diverse visual evidence. We therefore form a routing-query set $\mathcal{Q}_{r}=\mathcal{Q}_{\mathrm{ent}}\cup\{q_{\mathrm{glb}}\}$, where $\mathcal{Q}_{\mathrm{ent}}$ contains anchors of entity spans explicitly mentioned in the question, and $q_{\mathrm{glb}}$ is a global question query instantiated by the last non-punctuation question token. Under causal self-attention, this token provides a compact representation of the preceding question context. Entity queries localize explicit objects and regions, whereas the global question query captures complementary attribute, relation, and broader contextual requirements. By default, the anchor of an entity span is its last non-punctuation token. For each $q_j\in\mathcal{Q}_{r}$, we reconstruct its visual attention row from the MRoPE-transformed query and key states at layer $L$, and average the resulting probabilities across the $H$ attention heads:
\begin{equation}
    a^{L}_{j,i}
    =
    \frac{1}{H}
    \sum_{h=1}^{H} A^{L,h}_{q_j,i},
    \qquad a^{L}_{j,i}\geq 0 .
    \label{eq:head_aggregated_attention}
\end{equation}
The forward pass uses FlashAttention-2~\cite{dao2024flashattention}; this selective reconstruction is applied to the small routing-query set and does not materialize the full attention matrix. Since the visual attention mass $Z_j=\sum_{n=1}^{N}a^{L}_{j,n}$ is positive, we normalize the visual slice into an evidence distribution:
\begin{equation}
    p_j(i)
    =
    \frac{a^{L}_{j,i}}{Z_j},
    \qquad
    \sum_{i=1}^{N}p_j(i)=1.
    \label{eq:evidence_distribution}
\end{equation}
$p_j$ is computed after visual and textual tokens have interacted, and is therefore conditioned on the current question.

\subsection{4.2. Minimum-Coverage Evidence Routing}
\label{sec:independent_routing}

Averaging evidence distributions can suppress tokens that are vital to one key semantic aspect of the question. Thinking-Once routes evidence independently for each query. Let $[N]=\{1,\ldots,N\}$. For a coverage threshold $\rho\in(0,1]$, $S_j(\rho)$ is the canonical fixed-tie-breaking solution of
\begin{equation}
    S_j(\rho)
    \in
    \arg\min_{S\subseteq[N]} |S|
    \quad
    \mathrm{s.t.}\quad
    \sum_{i\in S}p_j(i)\geq\rho .
    \label{eq:min_coverage}
\end{equation}
This set is obtained by sorting $p_j(i)$ in descending order, using fixed tie-breaking, and taking the shortest valid prefix whose total mass reaches $\rho$. Focused queries retain few tokens, whereas diffuse queries automatically preserve more evidence. The final core set is $S_{\mathrm{core}}=\bigcup_{q_j\in\mathcal{Q}_{r}}S_j(\rho)$, which guarantees at least $\rho$ evidence coverage separately for every individual routing query. The union is not necessarily the global minimum-cardinality set satisfying all query constraints, but fixed tie-breaking makes $S_{\mathrm{core}}$ monotone in $\rho$.

\subsection{4.3. Context-Preserved Single-Visual-Pass Inference}
\label{sec:context_preserved_inference}

The core set preserves question-relevant tokens at full resolution, while the remaining visual tokens are summarized to retain coarse spatial contextual information. We partition the visual grid into disjoint cells $\mathcal{G}$ and define the unselected tokens in each cell as 
$U_g=g\cap([N]\setminus S_{\mathrm{core}})$. Each non-empty set is summarized by parameter-free mean pooling:
\begin{equation}
b_g
=
\frac{1}{|U_g|}
\sum_{i\in U_g}v_i^L,
\qquad |U_g|>0.
\label{eq:background_pooling}
\end{equation}
Let $\mathrm{OrdMerge}$ place retained core tokens and background tokens according to the original visual raster order. A background token is inserted at the position of the last token it summarizes, so routing does not expose future visual states to earlier routed positions. The routed visual evidence is
\begin{equation}
E^{L}
=
\mathrm{OrdMerge}\!\left(
V^{L}_{S_{\mathrm{core}}};
\{b_g:\ g\in\mathcal{G},\ |U_g|>0\}
\right).
\label{eq:evidence_representation}
\end{equation}
Replacing the original visual subsequence with $E^{L}$ while keeping the textual states unchanged produces the final routed multimodal hidden-state representation for decoding.
\begin{equation}
\widetilde{H}^{L}
=
\mathrm{SeqMerge}(E^{L};T^{L}).
\label{eq:routed_hidden_state}
\end{equation}
This construction preserves question-relevant evidence at full resolution while retaining coarse scene context. Under the decoder's causal mask, the raster and prompt order preserve the causal direction of information during routing.

\begin{table*}[!t]
\centering
\small
\setlength{\tabcolsep}{1.2pt}
\renewcommand{\arraystretch}{1}

% 第二层表头整体上移
\newcommand{\uphead}[1]{%
  \raisebox{0.60ex}[0pt][0pt]{#1}%
}

\begin{tabularx}{\textwidth}{
  @{}
  >{\raggedright\arraybackslash}p{3.0cm}
  >{\centering\arraybackslash}p{1.25cm}|
  *{3}{>{\centering\arraybackslash}X}|
  *{3}{>{\centering\arraybackslash}X}|
  *{3}{>{\centering\arraybackslash}X}|
  >{\centering\arraybackslash}X|
  >{\centering\arraybackslash}X
  @{}
}

\noalign{\hrule height 0.8pt}

\multirow{2}{*}{\textbf{Base Model}}
& \multirow{2}{*}{\textbf{Setting}}
& \multicolumn{3}{c|}{\textbf{V$^*$Bench}}
& \multicolumn{3}{c|}{\textbf{HRBench-4K}}
& \multicolumn{3}{c|}{\textbf{HRBench-8K}}
& \multirow{2}{*}{\textbf{Mean}}
& \multirow{2}{*}{
  \makecell[c]{\textbf{Mem.}\\[-1pt]\textbf{GB}$\downarrow$}
}
\\[-2pt]

\cmidrule(l{4pt}r{4pt}){3-5}
\cmidrule(l{4pt}r{4pt}){6-8}
\cmidrule(l{4pt}r{4pt}){9-11}

\noalign{\vskip -0.5pt}

& &
\uphead{Attr}
& \uphead{Spa.}
& \uphead{Avg}
& \uphead{FSP}
& \uphead{FCP}
& \uphead{Avg}
& \uphead{FSP}
& \uphead{FCP}
& \uphead{Avg}
& &
\\[-3pt]

\noalign{\hrule height 0.8pt}

\multirow{2}{*}{Qwen2.5-VL-7B}
& Base
& 80.9 & 76.3 & 79.1
& 86.3 & 55.5 & 70.9
& 83.5 & 51.3 & 67.4
& 72.5
& $\sim$20 \\

& Ours
& \textbf{89.6} & \textbf{88.2} & \textbf{89.0}
& \textbf{89.8} & \textbf{61.3} & \textbf{75.5}
& \textbf{87.5} & \textbf{58.3} & \textbf{72.9}
& \textbf{79.1}
& \textbf{$\sim$16} \\

\hline

\multirow{2}{*}{Qwen3-VL-8B}
& Base
& 84.4 & 80.3 & 82.7
& 91.3 & 65.8 & 78.5
& 84.0 & 62.8 & 73.4
& 78.2
& $\sim$20 \\

& Ours
& \textbf{87.8} & \textbf{81.6} & \textbf{85.3}
& \textbf{91.8} & \textbf{70.5} & \textbf{81.1}
& \textbf{84.3} & \textbf{64.3} & \textbf{74.3}
& \textbf{80.2}
& \textbf{$\sim$16} \\

\hline

\multirow{2}{*}{ZwZ-4B}
& Base
& 90.4 & 86.8 & 89.0
& 89.5 & 68.0 & 78.8
& 86.8 & 65.5 & 76.1
& 81.3
& $\sim$20 \\

& Ours
& \textbf{90.4} & \textbf{86.8} & \textbf{89.0}
& \textbf{90.0} & \textbf{73.0} & \textbf{81.5}
& \textbf{88.0} & \textbf{68.5} & \textbf{78.3}
& \textbf{82.9}
& \textbf{$\sim$16} \\

\hline

\multirow{2}{*}{ZwZ-7B}
& Base
& 90.4 & 84.2 & 88.0
& 90.0 & 59.5 & 74.8
& 88.5 & 58.3 & 73.4
& 78.7
& $\sim$20 \\

& Ours
& \textbf{90.4} & \textbf{89.5} & \textbf{90.1}
& \textbf{90.8} & \textbf{67.3} & \textbf{79.0}
& \textbf{89.8} & \textbf{63.5} & \textbf{76.6}
& \textbf{81.9}
& \textbf{$\sim$16} \\

\hline

\multirow{2}{*}{ZwZ-8B}
& Base
& 86.6 & 88.2 & 87.4
& 90.8 & 69.8 & 80.3
& 87.0 & 66.8 & 76.9
& 81.5
& $\sim$20 \\

& Ours
& \textbf{87.0} & \textbf{90.8} & \textbf{88.5}
& \textbf{91.8} & \textbf{70.5} & \textbf{81.1}
& \textbf{88.3} & \textbf{69.0} & \textbf{78.6}
& \textbf{82.7}
& \textbf{$\sim$16} \\

\hline

\multicolumn{2}{@{}l|}{\textbf{Average $\Delta$}}
& \textbf{+2.5} & \textbf{+4.2} & \textbf{+3.1}
& \textbf{+1.3} & \textbf{+4.8} & \textbf{+3.0}
& \textbf{+1.6} & \textbf{+3.8} & \textbf{+2.7}
& \textbf{+2.9}
& \textbf{-4} \\

\noalign{\hrule height 0.8pt}
\end{tabularx}

\caption{
Comprehensive results across five representative MLLMs.
Mean averages the three benchmark averages.
Average $\Delta$ is measured over the five base models against the
corresponding base setting; higher accuracy and lower memory cost are better.
}
\label{tab:comprehensive_main}
\end{table*}

We keep the MRoPE coordinates for retained visual tokens. Each background token takes the integer-rounded centroid of coordinates in \(U_g\). Let $M$ be the maximum coordinate across all MRoPE axes in the routed prompt. Generated text tokens then use the shared scalar position during decoding
\begin{equation}
\mathbf{p}_{\mathrm{gen},n}
=
(M+n,\ M+n,\ M+n),
\qquad n\geq1.
\label{eq:generation_position}
\end{equation}
Thus, routing shortens the KV cache of layers deeper than $L$ without losing the original spatial coordinate system or breaking generation continuity in later layers. Let $M_T$ be the number of text tokens and let $G_{\mathrm{act}}\leq G=|\mathcal{G}|$ be the number of non-empty background cells. The routed sequence length is $|\widetilde{H}^{L}|=M_T+|S_{\mathrm{core}}|+G_{\mathrm{act}}$. Since each background token replaces at least one unselected visual token,
\begin{equation}
|\widetilde{H}^{L}|
\leq
M_T+\min\!\left\{N,\ |S_{\mathrm{core}}|+G\right\}
\leq
M_T+N .
\label{eq:routed_length_bound}
\end{equation}
If $n_0=N+M_T$ and $n_1=|\widetilde{H}^{L}|$, the cost of each layer changes from $\mathcal{O}(n_0 d^2+n_0^2 d)$ to $\mathcal{O}(n_1 d^2+n_1^2 d)$. The routed length, the attention cost, and the KV cache decrease whenever at least one background token summarizes multiple unselected tokens. Thinking-Once thus treats efficiency primarily as a consequence of routing compact, question-conditioned evidence, rather than as generic token removal.

\section{5. Experiments}
\label{sec:experiments}

\subsection{5.1. Experimental Setup}
\label{sec:experimental_setup}

\paragraph{Models and Protocol.}
We evaluate Thinking-Once on multiple MLLMs, including Qwen2.5-VL-7B-Instruct, Qwen3-VL-8B-Instruct, and the ZwZ series~\cite{qwen2.5-vl,qwen3-vl,wei2026zooming}. Hereafter, we abbreviate the two instruction-tuned models as Qwen2.5-VL-7B and Qwen3-VL-8B. These models span different generations, parameter scales, and training recipes, allowing us to more broadly assess the generalizability of the proposed routing strategy across diverse models. All experiments follow the same single-visual-pass inference protocol, using one full-image visual encoding and one multimodal forward pass.

\paragraph{Benchmarks.}
We evaluate our method on three HR-VQA benchmarks:
V\textsuperscript{*}Bench~\cite{vstar2024}, HRBench-4K, and
HRBench-8K~\cite{hrbench2025}. V\textsuperscript{*}Bench measures fine-grained attribute and spatial perception, while HRBench evaluates single-instance perception (FSP) and cross-instance perception (FCP) under challenging high-resolution settings, following prior work \cite{liu2025hide,zheng2025deepeyes}.
Implementation details and results on MME-RealWorld-Lite~\cite{mme-realworld-2025},
ZoomBench~\cite{wei2026zooming}, TreeBench~\cite{TreeBench2026}, and
POPE~\cite{pope-2023} are provided in Appendices C and D.

\subsection{5.2. Main Results}
\label{sec:base_model_generalization}

Table~\ref{tab:comprehensive_main} evaluates Thinking-Once on five representative MLLM base models, including Qwen2.5-VL-7B~\cite{qwen2.5-vl}, Qwen3-VL-8B~\cite{qwen3-vl}, and ZwZ-series models~\cite{wei2026zooming}. Across these base models, Thinking-Once consistently improves or matches the corresponding Base setting on the three benchmark averages overall. On average, it improves V$^*$Bench by +3.1 points, HRBench-4K by +3.0 points, and HRBench-8K by +2.7 points, while also reducing memory usage by about 4\,GB. On the matched Qwen2.5-VL-7B base model, Thinking-Once improves V$^*$Bench by +9.9 points, HRBench-4K by +4.6 points, and HRBench-8K by +5.5 points, increasing the mean score from 72.5 to 79.1 while substantially reducing peak memory from about 20\,GB to about 16\,GB. With the stronger ZwZ-8B base model, Thinking-Once reaches a mean score of 82.7, indicating that the same internal evidence-routing principle remains broadly useful when the base model already has stronger high-resolution perception capability.

The gains are more pronounced on relation-sensitive metrics. On V$^*$Bench, Spatial accuracy improves by +4.2 points on average, compared with +2.5 points for Attribute accuracy. On HRBench, FCP improves by +4.8 points on 4K and +3.8 points on 8K, larger than the corresponding FSP gains of +1.3 and +1.6 points. This indicates that Thinking-Once is especially useful when the answer depends on relations, comparisons, and cross-region context, rather than only on isolated object attributes. The results therefore directly support our central view: HR-VQA failures are not only caused by missing visual evidence, but also by ineffective transmission of already encoded evidence to later reasoning layers.

\begin{table*}[!t]
\centering
\normalsize
\setlength{\tabcolsep}{1.35pt}
\renewcommand{\arraystretch}{1}

\begin{tabular*}{\textwidth}{
@{\extracolsep{\fill}}
lccc|ccc|ccc|ccc|cc@{}
}
\noalign{\hrule height 0.7pt}

\multirow{2}{*}{\textbf{Method}}
& \multirow{2}{*}{\makecell[c]{\textbf{Train.}\\\textbf{-free}}}
& \multirow{2}{*}{\makecell[c]{\textbf{Extra}\\\textbf{Visual}}}
& \multirow{2}{*}{\makecell[c]{\textbf{Time}\\\textbf{(min)}$\downarrow$}}
& \multicolumn{3}{c|}{\textbf{V$^*$Bench}}
& \multicolumn{3}{c|}{\textbf{HRBench-4K}}
& \multicolumn{3}{c|}{\textbf{HRBench-8K}}
& \multirow{2}{*}{\textbf{Mean}}
& \multirow{2}{*}{\makecell[c]{\textbf{Mem.}\\\textbf{GB}$\downarrow$}}
\\[-0.35ex]

\cmidrule(lr){5-7}
\cmidrule(lr){8-10}
\cmidrule(lr){11-13}

& & &
& Attr & Spa. & Avg
& FSP & FCP & Avg
& FSP & FCP & Avg
& & \\
[-0.20ex]

\noalign{\hrule height 0.8pt}

% \multicolumn{15}{@{}l}{\textit{Closed-source Models: reference results}} \\
% \hline

GPT-4o-1120 ~\cite{gpt4o}
& --
& --
& --
& -- & -- & 66.0
& 70.0 & 48.0 & 59.0
& 62.0 & 49.0 & 55.5
& 60.2
& -- \\

\hline
\multicolumn{15}{@{}l}{\textit{External Evidence Reacquisition: extra visual inputs or re-encoding}} \\
\hline

VLM-R$^3$~\cite{jiang2026vlm}
& \no
& \yes
& $\sim$30
& 86.1 & 71.1 & 80.1
& 81.0 & \underline{57.5} & 69.3
& 69.5 & 51.5 & 60.5
& 70.0
& $\sim$23 \\

DeepEyes~\cite{zheng2025deepeyes}
& \no
& \yes
& $\sim$37
& \underline{87.0} & 80.3 & 84.3
& 89.3 & 56.3 & 72.8
& 85.5 & 56.0 & 70.8
& 76.0
& $\sim$20 \\

TreeVGR~\cite{TreeBench2026}
& \no
& \yes
& \textbf{$\sim$5}
& 85.2 & 85.5 & \underline{85.3}
& 89.5 & 57.0 & 73.3
& 84.5 & \underline{57.3} & \underline{70.9}
& 76.5
& $\sim$18 \\

DyFo~\cite{li2025dyfo}
& \yes
& \yes
& $\sim$31
& 80.9 & \underline{86.8} & 83.3
& 89.3 & 54.0 & 71.6
& \underline{86.5} & 53.3 & 69.9
& 74.9
& $\sim$86$^\ddagger$ \\

DeepScan~\cite{DeepScan2026}
& \yes
& \yes
& $\sim$180
& \noindent\textbf{89.6} & \noindent\textbf{88.2} & \noindent\textbf{89.0}
& 89.0 & 58.0 & 73.5
& 85.0 & 56.5 & 70.8
& \underline{77.8}
& $\sim$40$^\ddagger$ \\

ViCrop~\cite{khayatkhoei2025mllms}
& \yes
& \yes
& $\sim$28
& \noindent\textbf{89.6} & 71.1 & 82.2
& \noindent\textbf{90.5} & \underline{57.5} & \underline{74.0}
& 85.5 & 53.0 & 69.3
& 75.2
& $\sim$20 \\

\hline
\multicolumn{15}{@{}l}{\textit{Compression-First Token Reduction: generic pruning or merging}} \\
\hline

BTP~\cite{chen2026btp}
& \yes
& \no
& $\sim$90
& 72.2 & 69.7 & 71.2
& 88.8 & 53.0 & 70.9
& 82.5 & 46.8 & 64.6
& 68.9
& \underline{$\sim$17} \\

TRIO~\cite{zhang2026trio}
& \yes
& \no
& $\sim$14
& 84.3 & 71.1 & 79.1
& 88.3 & 53.3 & 70.8
& 82.5 & 49.5 & 66.0
& 72.0
& $\sim$50$^\ddagger$ \\

HiPrune~\cite{liu2026hiprune}
& \yes
& \no
& $\sim$12
& 82.6 & 77.6 & 80.6
& 88.8 & 55.0 & 71.9
& 85.0 & 53.5 & 69.3
& 73.9
& $\sim$70$^\ddagger$ \\

VisionZip~\cite{yang2025visionzip}
& \yes
& \no
& $\sim$11
& 81.7 & 77.6 & 80.1
& 88.8 & 54.8 & 71.8
& 85.5 & 53.5 & 69.5
& 73.8
& $\sim$36$^\ddagger$ \\

V$^2$Drop~\cite{V2Drop2026}
& \yes
& \no
& \underline{$\sim$8}
& 80.0 & 77.6 & 79.1
& 88.0 & 54.8 & 71.4
& 84.8 & 50.5 & 67.6
& 72.7
& $\sim$20 \\

\hline
\multicolumn{15}{@{}l}{\textit{Ours: Single-Visual-Pass Evidence Routing without extra visual encoding}} \\
\hline

\textbf{Thinking-Once}
& \yes
& \no
& \noindent\textbf{$\sim$5}
& \noindent\textbf{89.6} & \noindent\textbf{88.2} & \noindent\textbf{89.0}
& \underline{89.8} & \noindent\textbf{61.3} & \noindent\textbf{75.5}
& \noindent\textbf{87.5} & \noindent\textbf{58.3} & \noindent\textbf{72.9}
& \noindent\textbf{79.1}
& \noindent\textbf{$\sim$16} \\

% \textbf{\textit{$\Delta$ (vs. Qwen2.5-VL-7B)}}
% &
% &
% &
% &
% \textit{+8.7} & \textit{+11.9} & \textit{+9.9}
% & \textit{+3.5} & \textit{+5.8} & \textit{+4.6}
% & \textit{+4.0} & \textit{+7.0} & \textit{+5.5}
% & \textit{+6.7}
% & \textit{-4} \\
% \hline
\textbf{Thinking-Once$^\dagger$}
& \yes
& \no
& $\sim$5
& 87.0 & 90.8 & 88.5
& 91.8 & 70.5 & 81.1
& 88.3 & 69.0 & 78.6
& 82.7
& $\sim$16 \\

\hline
\noalign{\hrule height 0.8pt}
\end{tabular*}

\caption{
Fine-grained comparison across HR-VQA paradigms. Open-source methods use Qwen2.5-VL-7B unless marked with $^\dagger$ (ZwZ-8B). Compression-first pruning or merging methods retain 33.3\% of visual tokens. Mean averages the three benchmarks; time and average peak memory are measured on V$^*$Bench. $^\ddagger$ denotes two-GPU inference for OOM cases, with memory summed across GPUs. Boldface and underlining indicate the best and second-best Qwen2.5-VL-7B-based results.
}
\label{tab:method_comparison}
\end{table*}

\begin{table}[t]
  \centering
  \small
  \setlength{\tabcolsep}{0.8pt}
  \renewcommand{\arraystretch}{0.95}

  \begin{tabular*}{\columnwidth}{
    @{}l|
    @{\extracolsep{\fill}}
    ccc|
    ccc|
    ccc|
    c@{}
  }
    \toprule
    \multirow{2}{*}{\textbf{Variant}}
    & \multicolumn{3}{c|}{\textbf{V$^*$}}
    & \multicolumn{3}{c|}{\textbf{HRBench-4K}}
    & \multicolumn{3}{c|}{\textbf{HRBench-8K}}
    & \multirow{2}{*}{\textbf{Mean}} \\
    \cmidrule(lr){2-4}
    \cmidrule(lr){5-7}
    \cmidrule(lr){8-10}
    & Attr & Spa. & Avg
    & FSP & FCP & Avg
    & FSP & FCP & Avg
    & \\
    \midrule
    % \hline

    Base
    & 80.9 & 76.3 & 79.1
    & 86.3 & 55.5 & 70.9
    & 83.5 & 51.3 & 67.4
    & 72.5 \\

    % \addlinespace[0.5pt]
    % \hline
    \midrule

    w/o bg.
    & \underline{82.6} & \underline{84.2} & \underline{83.3}
    & \underline{88.8} & \underline{59.5} & \underline{74.1}
    & \underline{83.8} & \underline{56.5} & \underline{70.1}
    & \underline{75.8} \\

    Denoise
    & 81.7 & 81.6 & 81.7
    & 88.3 & 57.8 & 73.0
    & 82.8 & 56.0 & 69.4
    & 74.7 \\

    Entity only
    & \underline{82.6} & \underline{84.2} & \underline{83.3}
    & \underline{88.8} & 59.0 & 73.9
    & \underline{83.8} & 56.0 & 69.9
    & 75.7 \\

    Global only
    & 81.7 & 75.0 & 79.1
    & 88.3 & 50.0 & 69.1
    & 82.8 & 48.3 & 65.5
    & 71.2 \\

    \midrule
    % \hline

    \noindent\textbf{Full}
    & \noindent\textbf{89.6} & \noindent\textbf{88.2} & \noindent\textbf{89.0}
    & \noindent\textbf{89.8} & \noindent\textbf{61.3} & \noindent\textbf{75.5}
    & \noindent\textbf{87.5} & \noindent\textbf{58.3} & \noindent\textbf{72.9}
    & \noindent\textbf{79.1} \\

    \bottomrule
  \end{tabular*}

  \caption{Ablations on Qwen2.5-VL-7B. Mean averages three benchmarks; all changes are relative to Full.
  }
  \label{tab:ablation_qwen25}
\end{table}

\subsection{5.3. Comparison with Existing Methods}
\label{sec:existing_method_comparison}

Table~\ref{tab:method_comparison} compares Thinking-Once with representative HR-VQA methods under the original high-resolution setting. The comparison shows two distinct limitations of existing paradigms. First, external evidence reacquisition methods can recover missing local details by constructing extra visual inputs, re-encoding selected regions, or performing multi-step visual search. This confirms that additional observation is useful when the required evidence is absent from the current representation. However, the gains are obtained by changing the visual input pipeline: methods such as DeepEyes~\cite{zheng2025deepeyes}, DeepScan~\cite{DeepScan2026}, and ViCrop~\cite{khayatkhoei2025mllms} introduce extra visual processing, and several runs require higher latency or memory. 
% These results suggest that reacquisition addresses an evidence-availability problem, but does not test whether evidence already encoded in the model can be preserved and used more effectively. 

Compression-first methods show that shortening the visual sequence is not sufficient for HR-VQA. BTP~\cite{chen2026btp}, TRIO~\cite{zhang2026trio}, HiPrune~\cite{liu2026hiprune}, VisionZip~\cite{yang2025visionzip}, and V$^2$Drop~\cite{V2Drop2026} avoid extra visual inputs, but their results are less stable under the same high-resolution setting. Several methods run out of memory with \texttt{max\_pixel}=16384 on a single NVIDIA A800 GPU, and the runnable compression baselines remain below the full Thinking-Once result. More importantly, the gap to Thinking-Once is most visible on spatial and cross-instance metrics, where the answer depends on relations among small entities and surrounding context. 
% This pattern indicates that many tokens that appear removable under a generic compression criterion still participate in the evidence structure needed for high-resolution reasoning.

Thinking-Once is complementary to both paradigms. Like token-reduction methods, it shortens the later-layer sequence, but it does not select tokens according to generic redundancy alone. Instead, it more directly and effectively routes a compact, question-conditioned evidence set from the model's own intermediate representation through independent per-query coverage and background summaries, so compression emerges as a consequence of evidence allocation. Unlike external evidence reacquisition, it requires no additional visual input or re-encoding.
The comparison thus supports a more specific conclusion: HR-VQA benefits when already encoded evidence is preserved and reallocated according to the current question, especially for relation-sensitive reasoning.

\subsection{5.4. Ablation Studies}
\label{sec:ablation}

Table~\ref{tab:ablation_qwen25} reports component ablations on
Qwen2.5-VL-7B, while the same removals on additional base models are provided in Appendix E. Every variant is worse than the full method, indicating that the gain comes from preserving a structured evidence set rather than from a single component. Removing background summaries lowers the mean score by 3.3 points, suggesting that entity tokens alone are not enough for many HR-VQA questions. Spatial relations, relative positions, and attribute comparisons require surrounding regions as contextual references. Denoising attention sinks causes an even larger 4.4-point drop, showing that non-core tokens cannot be treated as uniform noise; some may carry global context or aggregation signals needed by later layers. The query ablations show that entity queries and the global question query are complementary. Entity-only routing loses 3.4 points, while global-only routing loses 7.9 points. Precise entity anchors provide the main localization signal, whereas the global question query supplies broader attribute, relation, and contextual demand. The full method is best because it more effectively preserves both types of evidence.

\section{6. Conclusion}

We have shown that fine-grained evidence can remain available after initial visual encoding. Ineffective evidence allocation can cause errors when key signals are diluted by redundant background during later reasoning. Question-relevant entities become identifiable at intermediate layers, defining an evidence-routing window in which they can influence downstream computation. These core tokens require compact contextual support for relational reasoning and attribute comparison. Thinking-Once routes this structured evidence to deeper layers within a single visual pass, improving accuracy and efficiency across five MLLMs. The results support evidence routing as a practical complement to additional visual acquisition in utilization-limited HR-VQA settings.

\FloatBarrier

\bibliography{aaai2027}

\end{document}

\onecolumn
\appendix

\begin{center}
{\huge \bfseries Technical Appendix: \textit{Thinking-Once} Evidence Routing for High-Resolution Visual Question Answering}
\end{center}

\vspace{1em}

\begin{tcolorbox}[
    enhanced,
    colback=tocbg,
    colframe=tocborder,
    arc=3mm,
    boxrule=0.5pt,
    left=5mm, right=5mm, top=5mm, bottom=5mm,
    shadow={2mm}{-2mm}{0mm}{black!10}
]
    \normalsize
    \textbf{\Large Contents}
    \vspace{0.5em}
    \hrule height 0.5pt
    \vspace{0.8em}

    \begin{tabular}{@{} p{2.2cm} p{0.78\linewidth} @{}}

        \textbf{\color{black!60} Appendix A}
        & {\large \bfseries \sffamily Formal Properties of the Routing Operator}
        \dotfill Page \pageref{app:proofs} \\[-1em]
        & {\small \color{black!80}
        Detailed specification and formal properties of the evidence routing operator, including selective attention reconstruction, normalization, minimum-coverage routing, background pooling, positional handling, sequence-length bounds, causal ordering, and downstream computational cost.} \\[3em]

        \textbf{\color{black!60} Appendix B}
        & {\large \bfseries \sffamily Supporting Evidence from Prior Work}
        \dotfill Page \pageref{app:prior_evidence} \\[-1em]
        & {\small \color{black!80}
        Additional tables and figures from prior studies that support the analysis of visual evidence availability, interleaved images, tool use, and target-instance evidence.} \\[3em]

        \textbf{\color{black!60} Appendix C}
        & {\large \bfseries \sffamily Implementation and Evaluation Details}
        \dotfill Page \pageref{app:implementation} \\[-1em]
        & {\small \color{black!80}
        Evaluation protocol, benchmark descriptions, training-free setting, hardware setup, and efficiency reporting details.} \\[3em]

        \textbf{\color{black!60} Appendix D}
        & {\large \bfseries \sffamily Supplementary Evaluation Results}
        \dotfill Page \pageref{app:supplementary_results} \\[-1em]
        & {\small \color{black!80}
        Additional results on MME-RealWorld-Lite, ZoomBench, TreeBench, and POPE using the ZwZ-7B model.} \\[3em]

        \textbf{\color{black!60} Appendix E}
        & {\large \bfseries \sffamily Additional Experiments and Qualitative Examples}
        \dotfill Page \pageref{app:analysis} \\[-1em]
        & {\small \color{black!80}
        Controlled compression-first comparisons, component ablations, and layer-wise qualitative examples that visualize question-guided evidence localization.} \\[3em]

        \textbf{\color{black!60} Appendix F}
        & {\large \bfseries \sffamily Extended Analysis}
        \dotfill Page \pageref{app:extended_analysis} \\[-1em]
        & {\small \color{black!80}
        Cross-layer retention diagnostics, layer-wise object-attention enrichment, matched-budget evidence-coverage analysis, complementary evidence, applicability boundaries, and falsifiability discussion for the evidence-allocation account.} \\

    \end{tabular}
\end{tcolorbox}

\vspace{2em}

\section{Formal Properties of the Routing Operator}
\label{app:proofs}

\makeatletter
\@ifundefined{c@theorem}{}{
  \setcounter{theorem}{0}
  \renewcommand{\thetheorem}{A.\arabic{theorem}}
}
\@ifundefined{c@lemma}{}{
  \renewcommand{\thelemma}{A.\arabic{lemma}}
}
\@ifundefined{c@proposition}{}{
  \renewcommand{\theproposition}{A.\arabic{proposition}}
}
\@ifundefined{c@corollary}{}{
  \renewcommand{\thecorollary}{A.\arabic{corollary}}
}
\@ifundefined{c@remark}{}{
  \renewcommand{\theremark}{A.\arabic{remark}}
}
\makeatother

This section provides the detailed routing specification omitted from the main paper and establishes the formal properties of the routing operator used there. The guarantees concern selective attention reconstruction, normalization, minimum-cardinality coverage, per-query coverage, background pooling, sequence length, causal ordering at the routing operation, generation-position continuity, and downstream computational cost. They do not assume or claim that attention mass is a perfect measure of semantic sufficiency.

\subsection{Detailed Routing Specification and Preliminaries}

Let an MLLM contain $D$ transformer layers, and let $L<D$ be the routing layer. At layer $L$, the multimodal hidden state is
\begin{equation}
H^L=\mathrm{SeqMerge}(V^L;T^L),
\qquad
V^L=\{v_i^L\}_{i=1}^{N},
\label{app:eq:method_state}
\end{equation}
where $V^L$ and $T^L$ are the visual and textual hidden states, respectively, and $\mathrm{SeqMerge}$ follows their original prompt positions. Throughout this appendix, $H^L$ denotes the output hidden state produced after transformer layer $L$ has completed, whereas $A_{q_j,i}^{L,h}$ denotes the attention probability computed within layer $L$ from that layer's input states. Thinking-Once uses $A_{q_j,i}^{L,h}$ to select and summarize the output visual states $V^L$, and replaces only the visual subsequence after layer $L$; all textual states and their relative order are retained. Consequently, the first transformer layer affected by routing is layer $L+1$.

The routing-query set is
\begin{equation}
\mathcal{Q}_r=\mathcal{Q}_{\mathrm{ent}}\cup\{q_{\mathrm{glb}}\}.
\label{app:eq:routing_queries}
\end{equation}
Here, $\mathcal{Q}_{\mathrm{ent}}$ contains the last non-punctuation token of each extracted entity span, and $q_{\mathrm{glb}}$ is a global question query instantiated by the last non-punctuation question token. Under causal self-attention, this token provides a compact representation of the preceding question context. The entity queries provide explicit object- or region-level anchors, while the global question query captures complementary attribute, relation, and contextual demand.

For each routing query $q_j\in\mathcal{Q}_r$, we selectively reconstruct the corresponding visual attention row from the MRoPE-transformed query and key states. If $A_{q_j,i}^{L,h}$ denotes the attention probability from $q_j$ to visual token $i$ in head $h$, the head-averaged score is
\begin{equation}
a_{j,i}^{L}
=
\frac{1}{H}
\sum_{h=1}^{H}A_{q_j,i}^{L,h},
\qquad i\in[N].
\label{app:eq:head_aggregated_attention}
\end{equation}
The ordinary model forward pass continues to use FlashAttention-2~\cite{dao2024flashattention}. The reconstruction is applied only to the small set $\mathcal{Q}_r$ and therefore does not materialize the complete sequence-by-sequence attention matrix.

Throughout, $A_{q_j,i}^{L,h}$ denotes the exact causal-softmax probability. For each head $h$, the reconstruction computes the attention logits of $q_j$ against \emph{all} keys visible to $q_j$ under the causal mask, including the preceding textual keys, applies the softmax over this complete visible key set, and only then restricts the resulting probability row to the visual positions. Hence the full attention matrix is never materialized, yet the softmax normalizer of every reconstructed row is exact. The order of operations matters: because the full-row softmax denominators differ across heads, restricting each head to the visual slice and renormalizing \emph{before} head averaging would in general yield a distribution different from the one defined by Eqs.~\eqref{app:eq:head_aggregated_attention} and~\eqref{app:eq:evidence_distribution}. The visual-slice normalization in Eq.~\eqref{app:eq:evidence_distribution} is therefore applied exactly once, after averaging the exact per-head probabilities.

We assume a single-image, image-before-text prompt layout under full causal attention: all $N$ visual tokens form a visible visual prefix before every $q_j\in\mathcal{Q}_r$, and no visual position in this prefix is masked from a routing query. Multi-image and interleaved layouts are handled by restricting each routing query to its visible visual-token set; this generalization is made precise in Remark~\ref{rem:multi_image}.

Define the visual attention mass and its normalized evidence distribution as
\begin{equation}
Z_j=\sum_{i=1}^{N}a_{j,i}^{L},
\qquad
p_j(i)=\frac{a_{j,i}^{L}}{Z_j}.
\label{app:eq:evidence_distribution}
\end{equation}
Under the prompt-layout assumption, every visual token has a finite unmasked attention logit. Softmax therefore assigns positive probability to each visible visual token, head averaging preserves positivity, and $Z_j>0$.

Let $\pi_j$ be a permutation of $[N]=\{1,\ldots,N\}$ that sorts $p_j$ in non-increasing order. Ties are broken by a fixed deterministic rule: tokens with equal probability are ordered by ascending token index, i.e., by their original visual raster position. All statements below that invoke fixed tie-breaking refer to this rule. Define
\begin{equation}
C_j(k)=\sum_{t=1}^{k}p_j\!\left(\pi_j(t)\right),
\qquad C_j(0)=0,
\end{equation}
and, for $\rho\in(0,1]$,
\begin{equation}
k_j(\rho)=\min\{k\in[N]:C_j(k)\geq\rho\},
\qquad
S_j(\rho)=\{\pi_j(1),\ldots,\pi_j(k_j(\rho))\}.
\label{app:eq:canonical_set}
\end{equation}
The core set is
\begin{equation}
S_{\mathrm{core}}
=
\bigcup_{q_j\in\mathcal{Q}_r}S_j(\rho).
\label{app:eq:core_union}
\end{equation}
Operationally, Thinking-Once therefore (i) runs the unmodified sequence to layer $L$, (ii) constructs entity and global-question routing queries, (iii) reconstructs only their visual attention rows, (iv) selects a minimum-coverage set independently for each query, (v) preserves the union at full token resolution while summarizing the unselected background, and (vi) resumes the remaining layers with the routed sequence.

\subsection{Evidence Distribution and Minimum Coverage}

\begin{lemma}[Normalized evidence distribution]
\label{lem:normalization}
For every routing query with $Z_j>0$, $p_j$ is a probability distribution over $[N]$.
\begin{equation}
    p_j(i)\geq 0,
    \qquad
    \sum_{i=1}^{N}p_j(i)=1.
\end{equation}
\end{lemma}

\begin{proof}
Non-negativity follows from $a_{j,i}^{L}\geq0$ and $Z_j>0$. Summing Eq.~\eqref{app:eq:evidence_distribution} over $i$ gives $\sum_i p_j(i)=Z_j/Z_j=1$.
\end{proof}

\begin{proposition}[Minimum-cardinality coverage]
\label{prop:min_coverage}
For any $\rho\in(0,1]$, the canonical set $S_j(\rho)$ in Eq.~\eqref{app:eq:canonical_set} solves
\begin{equation}
    \min_{S\subseteq[N]}|S|
    \quad\mathrm{s.t.}\quad
    \sum_{i\in S}p_j(i)\geq\rho.
\end{equation}
\end{proposition}

\begin{proof}
The set is feasible because its mass is $C_j(k_j(\rho))\geq\rho$. For any set $S$ of cardinality $m$, the sum of its probabilities cannot exceed the sum of the $m$ largest probabilities:
\begin{equation}
    \sum_{i\in S}p_j(i)\leq C_j(m).
\end{equation}
If $m<k_j(\rho)$, the definition of $k_j(\rho)$ gives $C_j(m)<\rho$; hence no set with fewer than $k_j(\rho)$ elements is feasible. Therefore $S_j(\rho)$ has minimum cardinality. The ascending-index tie-breaking rule selects one canonical minimum-cardinality solution when several exist.
\end{proof}

\begin{corollary}[Nested routing and per-query coverage]
\label{cor:nested_coverage}
If $0<\rho_1\leq\rho_2\leq1$, then the canonical fixed-order solutions satisfy $S_j(\rho_1)\subseteq S_j(\rho_2)$. Moreover, for every $q_j\in\mathcal{Q}_r$,
\begin{equation}
    \sum_{i\in S_{\mathrm{core}}}p_j(i)\geq\rho,
\end{equation}
where $\rho$ is the threshold used to construct $S_{\mathrm{core}}$.
\end{corollary}

\begin{proof}
Because $C_j(k)$ is non-decreasing, reaching a larger threshold cannot require a shorter sorted prefix, so $k_j(\rho_1)\leq k_j(\rho_2)$. Both sets use the same fixed ordering $\pi_j$, which proves nesting for the canonical solution. Since $S_j(\rho)\subseteq S_{\mathrm{core}}$ and $p_j(i)\geq0$, the mass of the union for query $j$ is at least the mass of $S_j(\rho)$, which is at least $\rho$.
\end{proof}

\begin{remark}[Scope of the minimum guarantee]
Proposition~\ref{prop:min_coverage} establishes minimum cardinality separately for each query. The union $S_{\mathrm{core}}$ guarantees every query's coverage but is not claimed to be the globally smallest set satisfying all query constraints jointly.
\end{remark}

\begin{remark}[Multi-image and interleaved layouts]
\label{rem:multi_image}
For multi-image or interleaved prompts, let $\mathcal{V}_j\subseteq[N]$ denote the set of visual positions visible to $q_j$ under the attention mask; under the single-image, image-before-text assumption above, $\mathcal{V}_j=[N]$ for every routing query. In the general case, the sums in Eq.~\eqref{app:eq:evidence_distribution} range over $\mathcal{V}_j$ instead of $[N]$, so that $Z_j=\sum_{i\in\mathcal{V}_j}a_{j,i}^{L}$ and each $p_j$ is a probability distribution supported on $\mathcal{V}_j$; masked visual positions are not eligible for query $q_j$ and receive zero mass by convention. The minimum-coverage selection of Eq.~\eqref{app:eq:canonical_set} is then performed within $\mathcal{V}_j$, the core set remains the union of the per-query selections, and visual tokens visible to no routing query are treated as unselected background. All results transfer directly: Lemma~\ref{lem:normalization}, Proposition~\ref{prop:min_coverage}, and Corollary~\ref{cor:nested_coverage} hold verbatim with $[N]$ replaced by $\mathcal{V}_j$, while Propositions~\ref{prop:length_bound}--\ref{prop:position_continuity} and Corollary~\ref{cor:complexity} are unaffected because they depend only on the partition $\mathcal{G}$ and the physical raster positions, not on which queries selected the core tokens.
\end{remark}

\subsection{Context Preservation and Sequence-Length Bound}

This subsection formalizes the background-downsampled routing operator. The selected core evidence remains at full token resolution, while every unselected visual token contributes to a coarse background summary.

Let $\mathcal{G}$ be a disjoint partition of the visual-token grid. For each cell $g\in\mathcal{G}$, define
\begin{equation}
U_g
=
g\cap([N]\setminus S_{\mathrm{core}}).
\label{app:eq:background_cell}
\end{equation}
Only non-empty cells emit summaries. Let
\begin{equation}
\mathcal{G}_{\mathrm{act}}
=
\{g\in\mathcal{G}:|U_g|>0\},
\qquad
G_{\mathrm{act}}=|\mathcal{G}_{\mathrm{act}}|,
\qquad
G=|\mathcal{G}|.
\end{equation}
In our implementation, each active cell is summarized with parameter-free mean pooling:
\begin{equation}
b_g
=
\frac{1}{|U_g|}
\sum_{i\in U_g}v_i^L,
\qquad
g\in\mathcal{G}_{\mathrm{act}}.
\label{app:eq:bg_mean_pool}
\end{equation}
Mean pooling introduces no additional learned parameters and ensures that every unselected token contributes to exactly one background summary. The sequence-length and ordering results below require only that each active cell produces one token derived from its own source set $U_g$.

The routed visual evidence is
\begin{equation}
B^L
=
\{b_g:g\in\mathcal{G}_{\mathrm{act}}\},
\qquad
E^L
=
\mathrm{OrdMerge}\!\left(
\{v_i^L:i\in S_{\mathrm{core}}\};B^L
\right).
\label{app:eq:routed_visual_evidence}
\end{equation}
The routed multimodal hidden state is then
\begin{equation}
\widetilde{H}^{L}
=
\mathrm{SeqMerge}(E^L;T^L),
\label{app:eq:routed_hidden_state}
\end{equation}
where the original textual states and their relative prompt order are unchanged.

The operator $\mathrm{OrdMerge}$ orders retained core tokens and background summaries according to the original visual raster order. A summary corresponding to $U_g$ is assigned the representative insertion position
\begin{equation}
m_g=\max U_g.
\label{app:eq:summary_position}
\end{equation}
Thus, a summary is inserted no earlier than any original visual token that contributes to it. This physical insertion position is used for routed sequence order and is distinct from the MRoPE coordinate used for positional encoding.

\begin{proposition}[Routed sequence-length bound]
\label{prop:length_bound}
For $M_T$ textual tokens, the routed sequence length $n_1$ satisfies
\begin{equation}
n_1
=
M_T+|S_{\mathrm{core}}|+G_{\mathrm{act}}.
\label{app:eq:routed_length}
\end{equation}
Moreover,
\begin{equation}
n_1
\leq
M_T+\min\{N,\ |S_{\mathrm{core}}|+G\}
\leq
M_T+N.
\label{app:eq:length_bound}
\end{equation}
Furthermore, $n_1<M_T+N$ if and only if at least one active background cell summarizes more than one token.
\end{proposition}

\begin{proof}
The routed visual sequence contains exactly $|S_{\mathrm{core}}|$ retained core tokens and one background summary token for each active cell, so Eq.~\eqref{app:eq:routed_length} follows directly.

Since only non-empty cells emit summary tokens, we have $G_{\mathrm{act}}\leq G$, which gives
\begin{equation}
|S_{\mathrm{core}}|+G_{\mathrm{act}}
\leq
|S_{\mathrm{core}}|+G.
\end{equation}
This proves the second candidate in the minimum term.

Next, because $\mathcal{G}$ is a partition of $[N]$, the non-empty sets $U_g$ partition the unselected visual tokens $[N]\setminus S_{\mathrm{core}}$. Therefore,
\begin{equation}
\sum_{g\in\mathcal{G}_{\mathrm{act}}}|U_g|
=
N-|S_{\mathrm{core}}|.
\end{equation}
Each active cell is non-empty, so
\begin{equation}
G_{\mathrm{act}}
\leq
\sum_{g\in\mathcal{G}_{\mathrm{act}}}|U_g|
=
N-|S_{\mathrm{core}}|.
\end{equation}
Hence
\begin{equation}
|S_{\mathrm{core}}|+G_{\mathrm{act}}
\leq
N.
\end{equation}
Combining the two upper bounds gives Eq.~\eqref{app:eq:length_bound}.

Finally, equality $|S_{\mathrm{core}}|+G_{\mathrm{act}}=N$ holds exactly when $G_{\mathrm{act}}=\sum_{g\in\mathcal{G}_{\mathrm{act}}}|U_g|$, which happens exactly when every active $U_g$ is a singleton. Therefore the inequality is strict exactly when some active cell contains at least two unselected tokens, i.e., when at least one background summary merges multiple tokens.
\end{proof}

\begin{proposition}[Causal safety of ordered merging]
\label{prop:causal_safety}
Under a causal decoder mask, suppose retained core tokens preserve their original visual raster positions, each summary token of $U_g$ is inserted at $m_g=\max U_g$, and all textual states retain their original relative order. Then the routing operation does not place information from a later original sequence position into an earlier routed position.
\end{proposition}

\begin{proof}
For each routed visual token $e$, define its representative position $\operatorname{rep}(e)$ and source set $\operatorname{src}(e)$ as follows. If $e$ is a retained core token $v_i^L$, then $\operatorname{rep}(e)=i$ and $\operatorname{src}(e)=\{i\}$. If $e$ is a background summary token $b_g$, then $\operatorname{rep}(e)=m_g=\max U_g$ and $\operatorname{src}(e)=U_g$. Since the cells in $\mathcal{G}$ are disjoint, different active cells have disjoint $U_g$ sets. Moreover, $m_g\in U_g\subseteq[N]\setminus S_{\mathrm{core}}$, so a summary insertion position cannot coincide with a retained core-token position. Thus $\mathrm{OrdMerge}$ is well defined.

For a retained core token, clearly $\operatorname{src}(e)\subseteq\{1,\ldots,\operatorname{rep}(e)\}$. For a background summary token, every $i\in U_g$ satisfies $i\leq m_g$, and therefore
\[
\operatorname{src}(e)\subseteq\{1,\ldots,\operatorname{rep}(e)\}.
\]
Ordered merging sorts retained core tokens and background summaries by their representative positions, so the routed visual order is monotone in $\operatorname{rep}(e)$. Under a causal decoder mask, a routed state at representative position $\operatorname{rep}(e)$ can therefore depend only on source positions no later than $\operatorname{rep}(e)$. No information from a later original sequence position is moved to an earlier routed position, and no future-to-past information path is introduced by the routing operation.
\end{proof}

\begin{remark}
Proposition~\ref{prop:causal_safety} is deliberately limited to the routing operation. It does not claim that background downsampling preserves every pairwise attention relation or leaves the model output unchanged.
\end{remark}

\begin{proposition}[Generation-position continuity]
\label{prop:position_continuity}
The representative insertion position $m_g$ in Eq.~\eqref{app:eq:summary_position} is a physical raster-order position used for ordered merging and causal masking. It is distinct from the MRoPE coordinate assigned to the summary token for positional encoding. Retained visual tokens keep their original MRoPE coordinates. Each background summary token uses the integer-rounded centroid of the MRoPE coordinates of the tokens in $U_g$. Let $M$ be the maximum coordinate value over all MRoPE axes in the routed prompt. If generated token $n\geq1$ is assigned
\begin{equation}
\mathbf{p}_{\mathrm{gen},n}
=
(M+n)\mathbf{1}_3,
\end{equation}
then every generated coordinate is later than every routed-prompt coordinate on all three axes, and consecutive generated coordinates differ by one on each axis.
\end{proposition}

\begin{proof}
By definition, every routed-prompt coordinate component is at most $M$. For generated token $n\geq1$, every component of $\mathbf{p}_{\mathrm{gen},n}$ equals $M+n$, which is strictly larger than $M$. Moreover,
\begin{equation}
\mathbf{p}_{\mathrm{gen},n+1}
-
\mathbf{p}_{\mathrm{gen},n}
=
\mathbf{1}_3.
\end{equation}
Thus generation positions continue strictly and uniformly after the routed prompt.
\end{proof}

\begin{remark}[MRoPE coordinates of background summaries]
The centroid-based MRoPE coordinate of a background summary is used only for positional encoding, while causal safety is determined by the physical routed sequence order and the decoder mask. Therefore, possible non-monotonicity of summary MRoPE coordinates along the physical sequence does not affect the causal-order argument in Proposition~\ref{prop:causal_safety}. A separate consequence of integer rounding is that routed MRoPE coordinates need not be distinct: the rounded centroid of a background summary may coincide with the coordinate of a retained core token, or with that of another summary. Such coincidences are likewise harmless. MRoPE coordinates enter the computation only through relative rotary phases in the attention scores, which remain well defined when coordinates repeat, whereas sequence order, the causal mask, and KV-cache indexing are governed by the distinct physical insertion positions established in Proposition~\ref{prop:causal_safety}. Moreover, by Proposition~\ref{prop:position_continuity}, every generated-token coordinate is strictly larger than every routed-prompt coordinate, so any duplication is confined to the routed prompt and never involves generated positions.
\end{remark}

\subsection{Downstream Computational Cost}

\begin{corollary}[Non-increasing downstream cost]
\label{cor:complexity}
Let $n_0=M_T+N$ be the original prompt length at layers deeper than the routing layer, and let $n_1$ satisfy Eq.~\eqref{app:eq:length_bound}. For hidden dimension $d$, write the per-layer FLOPs of a standard Transformer layer as
\begin{equation}
F(n)=c_1nd^2+c_2n^2d,
\end{equation}
where $c_1,c_2>0$ are constants depending only on the architecture. Then the per-layer FLOPs change from $F(n_0)$ to $F(n_1)$, and the physical KV-cache length changes from $n_0$ to $n_1$. Both quantities are non-increasing, and they strictly decrease whenever $n_1<n_0$, i.e., whenever at least one active background cell summarizes multiple unselected tokens.
\end{corollary}

\begin{proof}
Proposition~\ref{prop:length_bound} gives $n_1\leq n_0$, with strict inequality when at least one active background cell summarizes multiple unselected tokens. For positive $d,c_1,c_2$, both $nd^2$ and $n^2d$ are strictly increasing in $n$, so $F(n)$ is strictly increasing in $n$. Therefore $F(n_1)\leq F(n_0)$, with strict inequality when $n_1<n_0$. KV-cache storage is linear in physical sequence length, so the same non-increasing and strict decrease statements hold for the physical KV-cache length.
\end{proof}

\begin{remark}[Scope of the efficiency guarantee]
The formal reduction applies to transformer layers deeper than $L$; the first $L$ layers still process the original sequence. Selective attention-row reconstruction, sorting, background pooling, and ordered merging introduce additional routing overhead. However, reconstruction is restricted to $|\mathcal{Q}_r|$ query rows rather than the full attention matrix, and the remaining operations are performed once at layer $L$. The resulting end-to-end latency and memory changes are therefore measured empirically in the main experiments rather than inferred solely from the asymptotic bound.
\end{remark}

\subsection{What the Formal Properties Do Not Establish}

The results above establish formal properties of the routing operator conditional on its attention-derived evidence distributions and background pooling rule. They show that the evidence distributions are normalized, the per-query minimum-coverage sets are well-defined, every routing query receives at least the prescribed coverage before union, background downsampling does not increase sequence length, ordered merging is causally safe under the stated insertion rule, generation positions remain continuous, and downstream physical sequence length, per-layer FLOPs, and KV-cache size are non-increasing.

These properties do not establish that $p_j(i)$ exactly measures causal importance, that a background summary token is semantically equivalent to all tokens it summarizes, that the routed representation is sufficient for every question, or that the answer distribution is unchanged by routing. Those claims require empirical validation, which is provided through controlled interventions, ablations, and benchmark results in the main paper and appendix.

\section{Supporting Evidence from Prior Work}
\label{app:prior_evidence}

\subsection{Recoverability of Visual Encoder Representations}
\label{app:longcat_evidence}

The main paper argues that high-resolution visual evidence does not always need to be reacquired through additional crops, zoom-in operations, or sub-image re-encoding. This argument relies on a necessary condition: the visual representations produced by the initial encoder should still retain recoverable information about the original image. Prior evidence from LongCat-Next
\cite{meituanlongcatteam2026longcatnextlexicalizingmodalitiesdiscrete} supports
this condition by showing that frozen visual encoder representations can preserve recoverable low-level and structural information.

\begin{figure}[!htbp]
    \centering
    \includegraphics[width=0.95\linewidth]{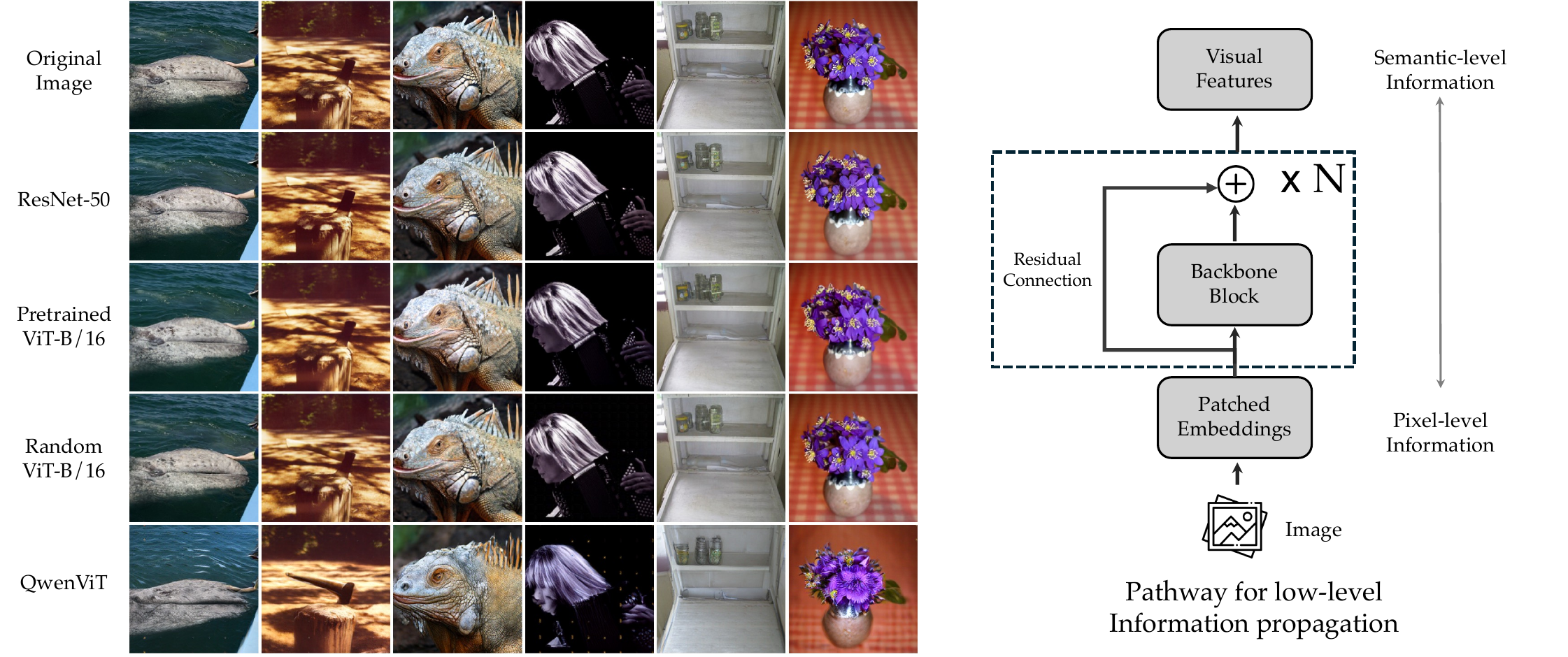}

\caption{
    Supporting evidence from prior work on visual reconstruction from frozen
    vision encoders. A lightweight pixel decoder can recover recognizable visual
    structures from different encoder outputs, suggesting that visual embeddings
    may preserve image-level information beyond high-level semantic abstraction.
    Figure adapted from LongCat-Next
    \cite{meituanlongcatteam2026longcatnextlexicalizingmodalitiesdiscrete}.
    }
    \label{fig:app_reconstruction}
\end{figure}

Figure~\ref{fig:app_reconstruction} shows visual reconstructions from several frozen vision encoders using a lightweight pixel decoder. Although the encoders are not trained specifically for pixel-level reconstruction in this analysis, the reconstructed images still preserve recognizable object contours, spatial layout, and local visual structures. This suggests that vision encoder outputs are not merely abstract semantic labels detached from the input image; instead, they may retain latent pathways through which fine-grained visual information survives the encoding process.

Table~\ref{tab:app_reconstruction_metrics} provides the corresponding quantitative reconstruction results. Across different visual encoder architectures, the reconstruction metrics indicate that a non-trivial amount of visual information remains recoverable from frozen encoder outputs. In particular, the strong reconstruction fidelity of the randomly initialized ViT-B/16 suggests that part of this recoverability may arise from architectural pathways, such as residual connections, rather than solely from semantic pretraining. The QwenViT result is especially relevant to MLLMs, as it suggests that the visual representations used by language-side reasoning can still carry recoverable image structure even after visual encoding.

\begin{table}[!htbp]
    \centering
    \scriptsize
    \setlength{\tabcolsep}{5.0pt}
    \renewcommand{\arraystretch}{0.96}
    \begin{tabular}{lcccc}
        \toprule
        \textbf{Metric}
        & \textbf{ResNet50}
        & \textbf{ViT-B/16}
        & \textbf{ViT-B/16}
        & \textbf{QwenViT} \\
        &
        & \textbf{(Pretrained)}
        & \textbf{(Random)}
        & \textbf{(w/o merger)} \\
        \midrule
        PSNR ($\uparrow$)
        & $20.88 \pm 3.44$
        & $21.86 \pm 3.14$
        & $30.52 \pm 3.42$
        & $18.16 \pm 2.61$ \\

        SSIM ($\uparrow$)
        & $0.509 \pm 0.174$
        & $0.581 \pm 0.139$
        & $0.887 \pm 0.051$
        & $0.46 \pm 0.14$ \\

        rFID ($\downarrow$)
        & 0.4619
        & 0.8850
        & 0.5847
        & 0.987 \\
        \bottomrule
    \end{tabular}

\caption{
    Quantitative reconstruction performance across visual encoder architectures.
    PSNR and SSIM measure reconstruction fidelity ($\uparrow$), while rFID
    measures perceptual discrepancy ($\downarrow$). Results are adapted from
    LongCat-Next
    \cite{meituanlongcatteam2026longcatnextlexicalizingmodalitiesdiscrete}.
    }
    \label{tab:app_reconstruction_metrics}
\end{table}

These prior results do not imply that every high-resolution question can be answered from the initial visual encoding, nor do they prove that the encoded representation is semantically sufficient for all downstream reasoning tasks. Rather, they support a more specific claim used in our analysis: fine-grained visual evidence can remain available after visual encoding. Therefore, when an MLLM fails on HR-VQA, the failure may not always be caused by the complete absence of visual evidence. It may also arise because the evidence that already exists in the visual embeddings or intermediate hidden states is diluted by irrelevant high-resolution tokens or is not effectively routed to later reasoning layers. This motivates Thinking-Once to operate on internal visual representations and route question-conditioned evidence within a single visual pass, instead of assuming that the model must always reacquire evidence through additional visual inputs.

\subsection{Interleaved Images Are Not Always Necessary}
\label{app:interleaved_evidence}

The main paper discusses that the gains of thinking-with-images or interleaved-image methods are not always attributable to the additional visual inputs themselves. The position paper by Yang et al.~\cite{yang2026position} provides direct supporting evidence by ablating interleaved images from several visual CoT models while keeping the evaluation protocol fixed. Their results, reproduced in Table~\ref{tab:app_interleaved_ablation}, show that removing interleaved images often leads to only small changes in performance, and in some benchmarks even slightly improves the score.

\begin{table}[!p]
\centering
\scriptsize
\setlength{\tabcolsep}{5.0pt}
\renewcommand{\arraystretch}{0.96}
\begin{tabular}{lcccc}
\toprule
\textbf{Method} & \textbf{V$^*$Bench} & \textbf{HRBench-4K} & \textbf{HRBench-8K} & \textbf{MME-Real-Lite} \\
\midrule
Qwen2.5-VL-7B & 76.4 & 68.1 & 65.5 & 44.5 \\
\midrule
DeepEyes-7B & 84.3 & 72.8 & 69.3 & 53.9 \\
\quad w/o Interleaved images
& 84.3 \neuchg{0.0}
& 72.3 \negchg{-0.5}
& 69.5 \poschg{+0.2}
& 53.5 \negchg{-0.4} \\
\midrule
Pixel-Reasoner-7B & 85.3 & 72.5 & 68.9 & 50.0 \\
\quad w/o Interleaved images
& 83.8 \negchg{-1.5}
& 72.5 \neuchg{0.0}
& 67.9 \negchg{-1.0}
& 49.3 \negchg{-0.7} \\
\midrule
Thyme-7B & 83.8 & 78.3 & 72.3 & 53.8 \\
\quad w/o Interleaved images
& 83.2 \negchg{-0.6}
& 78.5 \poschg{+0.2}
& 72.4 \poschg{+0.1}
& 53.9 \poschg{+0.1} \\
\midrule\midrule
\textbf{Method} & \textbf{ChartQA} & \textbf{OCRBench} & \textbf{LogicVista} & \textbf{MathVision} \\
\midrule
Qwen2.5-VL-7B & 86.1 & 88.2 & 46.1 & 26.6 \\
\midrule
DeepEyes-7B & 86.1 & 85.3 & 44.7 & 26.0 \\
\quad w/o Interleaved images
& 86.1 \neuchg{0.0}
& 85.0 \negchg{-0.3}
& 43.4 \negchg{-1.3}
& 26.6 \poschg{+0.6} \\
\midrule
Pixel-Reasoner-7B & 87.1 & 82.1 & 42.1 & 27.6 \\
\quad w/o Interleaved images
& 87.1 \neuchg{0.0}
& 82.5 \poschg{+0.4}
& 41.6 \negchg{-0.5}
& 27.3 \negchg{-0.3} \\
\midrule
Thyme-7B & 87.7 & 86.7 & 50.6 & 25.7 \\
\quad w/o Interleaved images
& 87.7 \neuchg{0.0}
& 86.8 \poschg{+0.1}
& 50.3 \negchg{-0.3}
& 25.7 \neuchg{0.0} \\
\bottomrule
\end{tabular}

\caption{
Ablation study of interleaved images within the visual CoT of
``Thinking with Images'' models. Values in parentheses indicate the performance
change after removing interleaved images relative to the corresponding full
model. The results are adapted from Yang et al.~\cite{yang2026position}.
}
\label{tab:app_interleaved_ablation}
\end{table}

The ablation results support a distinction between \emph{visual evidence reacquisition} and \emph{reasoning or utilization improvement}. If the interleaved images were consistently the dominant source of performance gains, removing them would be expected to cause large and systematic drops. Instead, the observed differences are often small, with several near-zero or positive changes. This suggests that part of the benefit of visual CoT or thinking-with-images training may come from improved language-side reasoning, task alignment, or better use of information already present in the original image representation, rather than solely from newly introduced visual observations.

These results should not be interpreted as evidence that interleaved images are never useful. Some settings, such as Pixel-Reasoner on V$^*$Bench and HRBench-8K, do show noticeable degradation after removing them. The more relevant conclusion for our work is narrower: additional images are not always necessary, and their presence is not always the primary explanation for improved performance. This motivates Thinking-Once to ask whether high-resolution visual evidence that has already been encoded and aligned inside the model can be routed more effectively within a single visual pass, before resorting to extra visual inputs.

\subsection{Tool Use Does Not Always Explain the Gain}
\label{app:tool_use_evidence}

Prior tool-use analyses provide complementary evidence that improvements from thinking-with-images or visual tool-use training may not come solely from the additional visual information obtained by tool calls. In particular, improvements can also arise from language-side reasoning alignment, better task adaptation, intrinsic capability changes, or reduced harmful tool interactions. This section summarizes two pieces of prior evidence that support our distinction between visual evidence reacquisition and visual evidence utilization.

Table~\ref{tab:app_imcot_ablation} reports an ablation from DeepEyes
\cite{zheng2025deepeyes}, where a model trained with text-only chain-of-thought
data is compared with the full interleaved multimodal chain-of-thought setting. Although the full iMCoT model achieves the strongest result on some benchmarks, text-only CoT training already brings a substantial improvement over the original Qwen2.5-VL-7B baseline on V$^*$Bench and HRBench-4K. Since text-only CoT does not introduce additional interleaved visual observations, this result suggests that part of the improvement of thinking-with-images systems may come from changes in reasoning behavior or task alignment, rather than from newly acquired visual inputs alone.

\begin{table}[!htbp]
    \centering
    \scriptsize
    \setlength{\tabcolsep}{5.0pt}
    \renewcommand{\arraystretch}{0.96}
    \begin{tabular}{lccc}
        \toprule
        \textbf{Model} & \textbf{V$^*$Bench} & \textbf{HRBench-4K} & \textbf{HRBench-8K} \\
        \midrule
        Qwen2.5-VL-7B
        & 71.2 & 68.8 & 65.3 \\
        RL w. Text-only CoT
        & 88.5 & 75.4 & 60.8 \\
        \midrule
        DeepEyes (iMCoT)
        & \textbf{90.1} & 75.1 & \textbf{72.6} \\
        \bottomrule
    \end{tabular}

\caption{
    Ablation on iMCoT from DeepEyes. Text-only CoT training already improves
    high-resolution VQA performance over the original Qwen2.5-VL-7B baseline,
    indicating that the benefit of thinking-with-images training is not entirely
    attributable to additional interleaved visual inputs. Results are adapted
    from DeepEyes \cite{zheng2025deepeyes}.
    }
    \label{tab:app_imcot_ablation}
\end{table}

\begin{figure}[!htbp]
    \centering
    \includegraphics[width=0.85\linewidth]{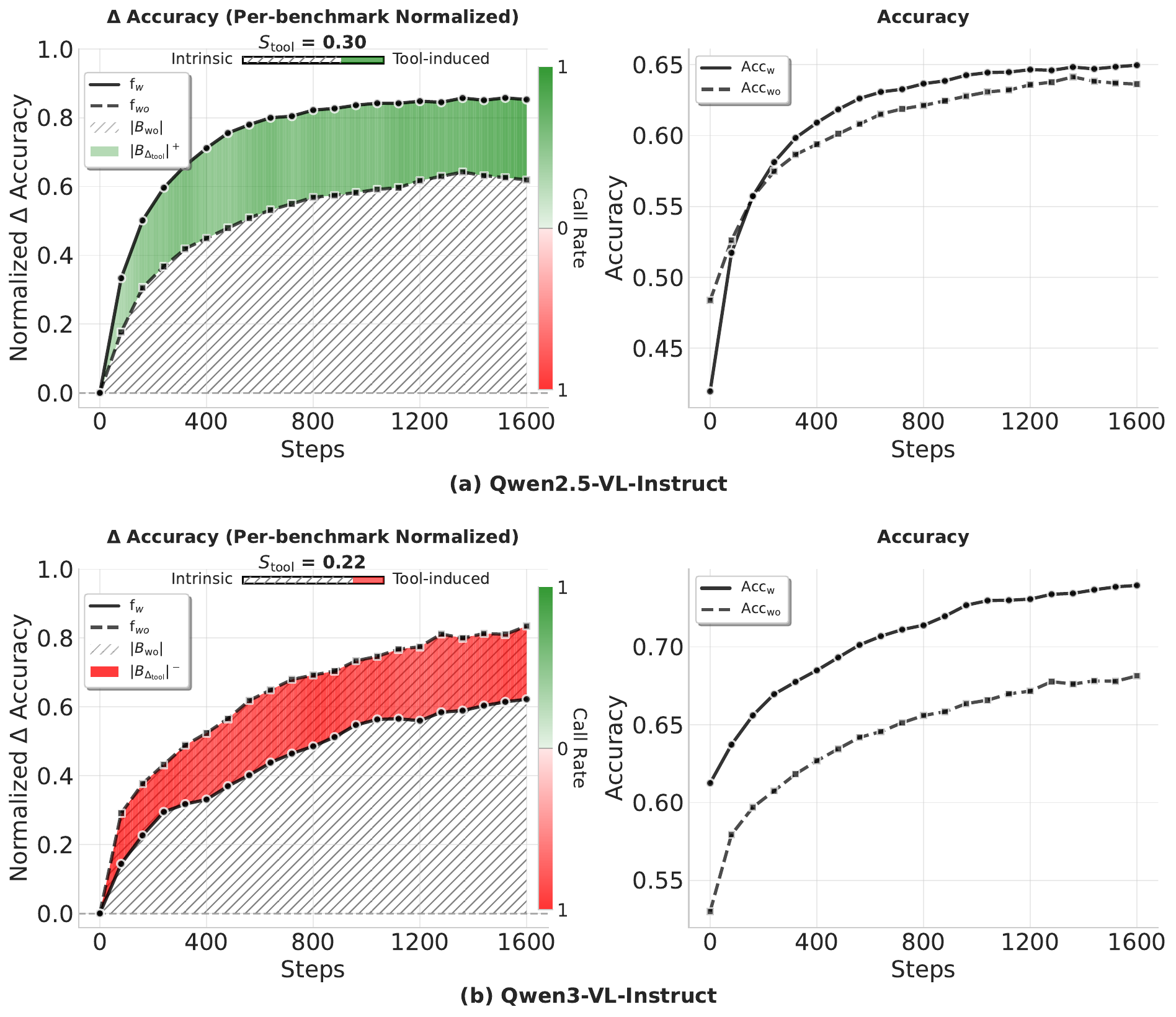}

\caption{
    Supporting evidence from the MED analysis of vision tool-use reinforcement
    learning. The analysis decomposes performance changes into intrinsic
    tool-free capability drift and tool-induced effects, showing that much of the
    observed gain can be attributed to intrinsic improvement rather than the
    additional crop-and-zoom tool effect alone. Figure adapted from
    Ma et al. \cite{ma2026does}.
    }
    \label{fig:app_tool_use_decomposition}
\end{figure}

A more direct decomposition is provided by Ma et al.
\cite{ma2026does}, which analyzes crop-and-zoom tool-use reinforcement learning
through the MED framework. Instead of only evaluating the final tool-available performance, the MED analysis evaluates training checkpoints under both tool-free and tool-available settings, and decomposes the observed performance change into tool-free intrinsic capability drift and tool-induced effects. Figure~\ref{fig:app_tool_use_decomposition} summarizes this analysis. The reported decomposition shows that a large fraction of the learning progress can be explained by tool-free intrinsic improvement, while the additional tool-induced component is comparatively smaller.

These two pieces of evidence are complementary. The DeepEyes ablation indicates that language-side training without additional interleaved images can already produce strong gains, while the MED analysis suggests that even tool-use reinforcement learning may improve performance partly through intrinsic model changes rather than through tool-induced visual reacquisition alone. These findings do not imply that interleaved images or crop-and-zoom tools are unnecessary in all cases. Instead, they show that attributing the gains of visual tool-use systems solely to newly acquired visual evidence is incomplete.

This distinction motivates the design of Thinking-Once. If a substantial part of the improvement in prior systems comes from better use of existing visual representations, improved reasoning behavior, or reduced interference, then high-resolution VQA should not be framed only as a problem of repeatedly acquiring new visual inputs. In the utilization-limited regime considered by Thinking-Once, the original full-image encoding and intermediate hidden states may already contain useful visual evidence, and the central problem is how to preserve, select, and route this evidence to later reasoning layers within a single visual pass.

\subsection{Target-Instance Evidence Is Necessary and Actionable}
\label{app:target_evidence}

The main paper argues that high-resolution visual reasoning depends on a small set of question-relevant visual evidence, and that these key evidence tokens should be preserved and routed rather than treated as interchangeable with generic background tokens. Prior work on traceable visual grounded reasoning provides supporting evidence for this view from the input-space perspective. In particular, TreeVGR \cite{TreeBench2026} studies whether annotated target instances are genuinely necessary for visual grounded reasoning through two complementary interventions: masking target instances and providing ground-truth bounding boxes as textual evidence hints.

Table~\ref{tab:app_treevgr_target_evidence} reports these two interventions on TreeBench. In the target-masking setting, removing all annotated target instances causes a consistent and substantial performance drop across different MLLMs. This indicates that the target instances are not merely high-quality annotations or convenient visual explanations, but contain indispensable evidence for answering the questions. In other words, the visual information associated with these target instances cannot be reliably replaced by the remaining background or global image context.

Complementarily, the textual-box-hint setting shows that explicitly providing ground-truth bounding boxes as evidence hints improves performance for all evaluated models. This result suggests that making the target evidence explicit helps models focus their reasoning on the relevant visual regions and reduces reliance on global image impressions or language priors. Together, the two interventions provide both a necessity and usefulness signal: removing target evidence hurts performance, while explicitly identifying it improves performance.

\begin{table}[!htbp]
    \centering
    \scriptsize
    \setlength{\tabcolsep}{2.8pt}
    \renewcommand{\arraystretch}{0.96}

    \textbf{Panel A: Masked target instances}\\[0.3em]
    \resizebox{0.95\textwidth}{!}{
    \begin{tabular}{c llllll}
        \toprule
        \textbf{Masking}
        & \textbf{Qwen2.5-VL-7B}
        & \textbf{InternVL3-8B}
        & \textbf{GPT-4o}
        & \textbf{o3}
        & \textbf{Gemini-2.5-Flash}
        & \textbf{Gemini-2.5-Pro} \\
        \midrule
        --
        & 37.0 & 38.8 & 46.9 & 54.8 & 45.9 & 54.1 \\
        \cmark
        & $31.8\,(\negnum{\downarrow 5.2})$
        & $29.6\,(\negnum{\downarrow 9.2})$
        & $29.1\,(\negnum{\downarrow 17.8})$
        & $33.8\,(\negnum{\downarrow 21.0})$
        & $29.9\,(\negnum{\downarrow 16.0})$
        & $33.1\,(\negnum{\downarrow 21.0})$ \\
        \bottomrule
    \end{tabular}}

    \vspace{0.8em}

    \textbf{Panel B: Explicit bounding-box-based textual hints}\\[0.3em]
    \resizebox{0.95\textwidth}{!}{
    \begin{tabular}{c llllll}
        \toprule
        \textbf{Textual Boxes}
        & \textbf{Qwen2.5-VL-7B}
        & \textbf{InternVL3-8B}
        & \textbf{GPT-4o}
        & \textbf{o3}
        & \textbf{Gemini-2.5-Flash}
        & \textbf{Gemini-2.5-Pro} \\
        \midrule
        --
        & 37.0 & 38.8 & 46.9 & 54.8 & 45.9 & 54.1 \\
        \cmark
        & $43.7\,(\posnum{\uparrow 6.7})$
        & $43.5\,(\posnum{\uparrow 4.7})$
        & $49.4\,(\posnum{\uparrow 2.5})$
        & $58.3\,(\posnum{\uparrow 3.5})$
        & $51.9\,(\posnum{\uparrow 6.0})$
        & $61.0\,(\posnum{\uparrow 6.9})$ \\
        \bottomrule
    \end{tabular}}

\caption{
    Target-instance evidence on TreeBench. Panel A shows that masking all
    annotated target instances causes consistent performance degradation across
    models. Panel B shows that providing ground-truth target boxes as textual
    evidence hints produces consistent gains. Results are adapted from TreeVGR
    \cite{TreeBench2026}.
    }
    \label{tab:app_treevgr_target_evidence}
\end{table}

These results support the central assumption behind our evidence-routing analysis. Although TreeVGR operates at the level of annotated input-space instances, the conclusion naturally connects to token-level routing in MLLMs: visual tokens corresponding to question-relevant target instances are likely to carry high-value reasoning evidence. If such instances are removed, the remaining visual context cannot reliably compensate for the missing evidence; if they are explicitly identified, model reasoning becomes more accurate.

This prior evidence therefore complements our layer-wise oracle intervention in the main paper. The oracle intervention shows that target-region tokens are highly informative inside intermediate model representations, while TreeVGR shows that the corresponding target instances are necessary and actionable in visual grounded reasoning. Together, they motivate Thinking-Once to preserve and route question-relevant core evidence tokens within a single visual pass, rather than applying query-agnostic compression that may discard the visual evidence most responsible for the answer.

\section{Implementation and Evaluation Details}
\label{app:implementation}

We provide additional implementation details for Thinking-Once Evidence Routing. Unless otherwise specified, all methods are evaluated under the same high-resolution visual question answering setting.

\paragraph{Evaluation protocol.}
All methods are evaluated on high-resolution visual question answering benchmarks, including V$^*$Bench, HRBench-4K, and HRBench-8K. For V$^*$Bench, we report attribute accuracy, spatial accuracy, average accuracy, inference time, and GPU memory usage. For HRBench-4K and HRBench-8K, we report FSP, FCP, and average accuracy. Additional supplementary evaluations are conducted on MME-RealWorld-Lite, ZoomBench, TreeBench, and POPE to examine real-world high-resolution perception, fine-grained zoom-oriented perception, traceable visual grounded reasoning, and object hallucination robustness.

\subsection{Benchmark and Dataset Details}
\label{app:dataset_details}

\paragraph{V$^*$Bench.}
V$^*$Bench is a high-resolution visual question answering benchmark designed to test whether MLLMs can identify and reason over fine-grained visual evidence in crowded images~\cite{vstar2024}. In our evaluation, it contains 191 multiple-choice questions, including 115 direct-attribute questions and 76 relative-position questions. The attribute split mainly evaluates whether a model can recognize localized visual properties of small or subtle objects, while the spatial split focuses on relative position and relation judgment. We report Attribute, Spatial, and their average accuracy.

\paragraph{HRBench-4K and HRBench-8K.}
HRBench is constructed for high-resolution multimodal perception and contains two complementary versions, HRBench-8K and HRBench-4K~\cite{hrbench2025}. The 8K version preserves high-resolution images with an average resolution around 8K, while the 4K version is obtained by cropping question-relevant target regions from the corresponding 8K images using human annotations. Both versions contain two sub-tasks: Fine-grained Single-instance Perception (FSP), which evaluates localized attributes, OCR, and visual prompting around a single target, and Fine-grained Cross-instance Perception (FCP), which evaluates cross-object or cross-region reasoning such as spatial relationships, map analysis, and chart analysis. We report FSP, FCP, and their average accuracy for both HRBench-4K and HRBench-8K.

\paragraph{MME-RealWorld-Lite.}
MME-RealWorld is a manually annotated benchmark for evaluating MLLMs in high-resolution real-world scenarios~\cite{mme-realworld-2025}. The full benchmark contains diverse real-world images and question-answer pairs covering multiple practical scenarios and subtasks. We use its lightweight evaluation subset, MME-RealWorld-Lite, for supplementary evaluation. It measures whether a model can perceive fine details and reason over realistic high-resolution content under practical visual conditions. We report Perception, Reasoning, and Overall scores.

\paragraph{ZoomBench.}
ZoomBench is a fine-grained multimodal perception benchmark introduced together with the Zooming-without-Zooming framework~\cite{wei2026zooming}. It contains 845 hybrid-annotated VQA examples spanning six fine-grained perceptual dimensions and is designed to measure whether a model can recognize small, localized, or detail-sensitive evidence that often motivates zoom-in operations. The benchmark also supports a dual-view protocol for analyzing the gap between global-image understanding and regional visual evidence. We report its multiple-choice score, blank-answer score, and overall score.

\paragraph{TreeBench.}
TreeBench is a diagnostic benchmark for traceable visual grounded reasoning, introduced by TreeVGR~\cite{TreeBench2026}. It focuses on subtle targets in complex scenes, requires evidence to be traceable through bounding-box annotations, and emphasizes second-order reasoning over object interactions and spatial hierarchies. The benchmark contains 405 challenging VQA pairs annotated by LMM experts. We use TreeBench to evaluate whether evidence routing helps not only answer prediction, but also reasoning over visually grounded and interaction-dependent evidence. We report Perception, Reasoning, and Overall scores.

\paragraph{POPE.}
POPE is a polling-based benchmark for evaluating object hallucination in large vision-language models~\cite{pope-2023}. It converts hallucination evaluation into binary object-existence questions, such as whether a queried object is present in the image. The benchmark constructs negative object queries under three settings: Random, Popular, and Adversarial. These splits test whether a model falsely predicts non-existent objects under increasingly difficult negative sampling strategies. We report the Random, Popular, Adversarial, and average scores.

\begin{table}[!htbp]
\centering
\scriptsize
\setlength{\tabcolsep}{3.0pt}
\renewcommand{\arraystretch}{0.96}
\resizebox{0.95\textwidth}{!}{
\begin{tabular}{l l l l}
\toprule
\textbf{Benchmark} & \textbf{Main focus} & \textbf{Reported metrics} & \textbf{Evaluation role} \\
\midrule
V$^*$Bench & Attribute and spatial reasoning in high-resolution scenes
& Attr / Spatial / Avg & Main benchmark \\
HRBench-4K & Cropped 4K fine-grained perception from question-relevant regions
& FSP / FCP / Avg & Main benchmark \\
HRBench-8K & Full high-resolution 8K perception and reasoning
& FSP / FCP / Avg & Main benchmark \\
MME-RealWorld-Lite & Real-world high-resolution perception and reasoning
& Perception / Reasoning / Overall & Supplementary benchmark \\
ZoomBench & Fine-grained zoom-oriented visual perception
& MCQ / Blank / Overall & Supplementary benchmark \\
TreeBench & Traceable visual grounded reasoning with evidence boxes
& Perception / Reasoning / Overall & Supplementary benchmark \\
POPE & Object hallucination under object-existence probing
& Random / Popular / Adversarial / Avg & Supplementary benchmark \\
\bottomrule
\end{tabular}}

\caption{
Summary of the evaluation benchmarks used in the main paper and appendix.
}
\label{tab:app_dataset_summary}
\end{table}

\paragraph{Inference settings.}
For background preservation, the visual-token map is uniformly partitioned into an $8\times8$ grid, and the unselected tokens within each non-empty cell are summarized into one background token through mean pooling.

\paragraph{Training-free setting and inference stages.}
Thinking-Once Evidence Routing is a training-free inference-time method and requires no additional supervised fine-tuning on the target benchmarks. The term \emph{single-visual-pass} refers specifically to processing and encoding the input image only once; it does not mean that the complete pipeline performs only one model invocation. For each sample, the pipeline sequentially invokes the same MLLM twice. The first invocation operates in text-only mode to extract question-relevant entity mentions from the cleaned question. The second invocation performs multimodal inference, during which the original full-resolution image is encoded once, intermediate-layer evidence routing is applied, and the final answer is generated. Entity extraction therefore introduces an additional text-only model call, but no additional visual input, crop, or image encoding.

\paragraph{Routing query construction.}
For entity-side routing, the same MLLM used for multimodal routing and answer generation is prompted in text-only mode to extract question-relevant entity mentions. Entity extraction is performed independently for each sample, without batching or caching, and extracted entities are not reused across samples. The resulting mentions are used only to construct internal routing queries; they do not introduce additional visual inputs, crops, or image encodings. Before extraction, we remove the dataset-specific answer-format suffix so that the text-only prompt focuses on the semantic content of the question rather than the multiple-choice output instruction.

\begin{small}
\begin{verbatim}
ENTITY_PROMPT_TEMPLATE = 
You are a highly precise language analysis engine. Your sole function is to extract entities
(e.g., objects, people) from a user's question.
Return a single line wrapped in <FINAL_OUTPUT> and </FINAL_OUTPUT>.
Lowercase only. Separate multiple entities with a comma and a space.
Question:

Answer Instruction = 
Answer with the option's letter from the given choices directly.
\end{verbatim}
\end{small}

\paragraph{Solution for empty entity extraction.}
If the text-only extractor returns an empty string or the designated no-entity
response, we do not disable evidence routing or revert to the unmodified base
model. Instead, we deterministically derive fallback entity phrases from the
cleaned question stem. For relation templates, we retain the two compared
mentions; for attribute templates, we retain the referenced object phrase; and
when no template matches, we use the complete cleaned question stem. The
resulting phrases are used as entity-side routing queries, while the global
question query, instantiated by the last non-punctuation question token, is
retained as usual. The same independent minimum-coverage selection and
background summarization are then applied without an additional model call,
visual encoding, or learned fallback component. Consequently, an empty
extraction degrades to deterministic question-derived routing rather than
unconditioned pruning. The \emph{Global only} setting reported in the ablation
tables is a deliberately constructed component ablation and is not the default
failure-handling path.

\paragraph{Routed token ratio.}
Table~\ref{tab:app_token_ratio} reports the routed visual-token ratios of Thinking-Once on the five base models used in the main experiments. The reported values exclude ablation variants and measure the proportion of visual tokens kept after evidence routing.

\begin{table}[!htbp]
\centering
\scriptsize
\setlength{\tabcolsep}{5.0pt}
\renewcommand{\arraystretch}{0.96}
\begin{tabular}{lcccc}
\toprule
\textbf{Base Model}
& \textbf{V$^*$Bench}
& \textbf{HRBench-4K}
& \textbf{HRBench-8K}
& \textbf{Avg.} \\
\midrule
Qwen2.5-VL-7B & 28.37 & 14.39 & 22.49 & 21.75 \\
ZwZ-7B        & 30.52 & 16.16 & 16.93 & 21.20 \\
Qwen3-VL-8B   & 23.60 & 30.05 & 27.31 & 26.99 \\
ZwZ-4B        & 18.76 & 27.39 & 23.54 & 23.23 \\
ZwZ-8B        &  6.18 & 13.43 & 21.96 & 13.86 \\
\midrule
Average       & 21.49 & 20.28 & 22.45 & 21.41 \\
\bottomrule
\end{tabular}

\caption{
Routed visual-token ratios of Thinking-Once. Values indicate the retained
visual-token ratio (\%) after evidence routing; ablation variants are excluded.
}
\label{tab:app_token_ratio}
\end{table}

The average routed ratio is $21.41\%$, meaning that Thinking-Once forwards only about one fifth of the visual-token sequence to the later layers after routing. The benchmark-level averages are also close, ranging from $20.28\%$ on HRBench-4K to $22.45\%$ on HRBench-8K, which suggests that the reduction is not specific to a single benchmark. At the same time, the ratio varies across MLLMs, from $13.86\%$ on ZwZ-8B to $26.99\%$ on Qwen3-VL-8B. This adaptive behavior is consistent with our routing objective: the method does not enforce a fixed compression rate, but preserves the amount of core evidence and compact background context required by the current model and input.

\paragraph{Hardware and efficiency reporting.}
All benchmark experiments are conducted on NVIDIA A800 GPUs. Efficiency results are reported on V$^*$Bench, with inference time measured in minutes and GPU memory measured in \,GB; lower values indicate better efficiency. The reported inference time is the end-to-end wall-clock time of the complete evaluation pipeline. For Thinking-Once, it includes both the unbatched and uncached text-only entity-extraction invocation and the subsequent multimodal routing-and-answer invocation, which are executed sequentially using the same MLLM. Thus, the approximately five-minute runtime reported for Thinking-Once represents entity extraction plus multimodal inference over the complete V$^*$Bench, rather than the cost of a single model invocation. Whenever possible, time and memory are measured on a single A800 GPU. For methods that encounter out-of-memory errors on a single A800 but can run with two A800 GPUs, we report measurements from the two-GPU execution. Memory denotes the dataset-averaged peak GPU usage; for a two-GPU run, we first average the peak usage of each GPU over V$^*$Bench and then sum the two GPU-wise averages.

\section{Supplementary Evaluation Results}
\label{app:supplementary_results}

This section reports supplementary evaluation results beyond the main V$^*$Bench, HRBench-4K, and HRBench-8K comparisons. Table~\ref{tab:app_additional_zwz7b_results} shows results on MME-RealWorld-Lite, ZoomBench, TreeBench, and POPE using the ZwZ-7B model. These benchmarks complement the main evaluation by covering real-world high-resolution perception, zoom-oriented fine-grained recognition, traceable visual grounded reasoning, and object hallucination robustness.

\begin{table}[!htbp]
\centering
\scriptsize
\setlength{\tabcolsep}{3.2pt}
\renewcommand{\arraystretch}{1.05}
\resizebox{\textwidth}{!}{%
\begin{tabular}{@{}lccccccccccccc@{}}
\toprule
\multirow{2}{*}{\textbf{Setting}}
& \multicolumn{3}{c}{\textbf{MME-RealWorld-Lite}}
& \multicolumn{3}{c}{\textbf{ZoomBench}}
& \multicolumn{3}{c}{\textbf{TreeBench}}
& \multicolumn{4}{c}{\textbf{POPE}} \\
\cmidrule(lr){2-4}
\cmidrule(lr){5-7}
\cmidrule(lr){8-10}
\cmidrule(lr){11-14}
& Perception & Reasoning & Overall
& MCQ & Blank & Overall
& Perception & Reasoning & Overall
& Random & Popular & Adv. & Avg \\
\midrule
GPT-4o
& 49.1 & 42.1 & 46.4
& -- & -- & --
& -- & -- & --
& -- & -- & -- & -- \\

Base
& 56.8 & 42.7 & 51.3
& 61.9 & 34.8 & 54.7
& 27.0 & 34.8 & 31.9
& 88.6 & 87.7 & 87.0 & 87.8 \\

ViCrop
& 57.8 & 42.1 & 51.7
& 57.6 & 35.3 & 51.7
& 28.9 & 33.2 & 31.6
& 86.3 & 85.4 & 85.1 & 85.6 \\

\textbf{Ours}
& \textbf{59.2} & \textbf{43.9} & \textbf{53.2}
& \textbf{62.9} & \textbf{35.7} & \textbf{55.7}
& \textbf{30.4} & \textbf{35.2} & \textbf{33.4}
& \textbf{90.4} & \textbf{88.6} & \textbf{87.7} & \textbf{88.9} \\

$\Delta$
& \posnum{+2.4} & \posnum{+1.2} & \posnum{+1.9}
& \posnum{+1.0} & \posnum{+0.9} & \posnum{+1.0}
& \posnum{+3.4} & \posnum{+0.4} & \posnum{+1.5}
& \posnum{+1.8} & \posnum{+0.9} & \posnum{+0.7} & \posnum{+1.1} \\
\bottomrule
\end{tabular}%
}
\caption{
Additional benchmark results on ZwZ-7B. We report MME-RealWorld-Lite,
ZoomBench, TreeBench, and POPE, with all scores rounded to one decimal place.
$\Delta$ denotes the absolute percentage-point improvement over Base. Higher
values indicate better performance.
}
\label{tab:app_additional_zwz7b_results}
\end{table}

The results show that Thinking-Once consistently improves the ZwZ-7B model across all four additional benchmarks. The gains are especially visible on MME-RealWorld-Lite and TreeBench, suggesting that evidence routing benefits both general real-world visual reasoning and traceable grounded reasoning. The improvement on POPE further indicates that routing question-relevant evidence does not increase object hallucination; instead, it slightly improves robustness across random, popular, and adversarial splits.

\section{Additional Experiments and Qualitative Examples}
\label{app:analysis}

\subsection{Additional Experiments}

\begin{table*}[!t]
\centering
\normalsize
\setlength{\tabcolsep}{3.0pt}
\renewcommand{\arraystretch}{1.0}

\begin{tabular*}{\textwidth}{
@{\extracolsep{\fill}}
lc|
ccc|
ccc|
ccc|
ccc
@{}
}
\noalign{\hrule height 0.8pt}

\multirow{2}{*}{\textbf{Base Model}}
& \multirow{2}{*}{\textbf{Setting}}
& \multicolumn{3}{c|}{\textbf{V$^*$Bench}}
& \multicolumn{3}{c|}{\textbf{HRBench-4K}}
& \multicolumn{3}{c|}{\textbf{HRBench-8K}}
& \multirow{2}{*}{\textbf{Mean}}
& \multirow{2}{*}{
  \makecell[c]{\textbf{Time}\\\textbf{(min)}$\downarrow$}
}
& \multirow{2}{*}{
  \makecell[c]{\textbf{Mem.}\\\textbf{GB}$\downarrow$}
} \\
\cline{3-5}
\cline{6-8}
\cline{9-11}

&
& Attr & Spa. & Avg
& FSP & FCP & Avg
& FSP & FCP & Avg
& & & \\

\noalign{\hrule height 0.8pt}

\multirow{4}{*}{\textbf{ZwZ-7B}}
& Base
& 90.4 & 84.2 & 88.0
& 90.0 & 59.5 & 74.8
& 88.5 & 58.3 & 73.4
& 78.7
& \textbf{$\sim$5}
& $\sim$20 \\

& HiDe$^\ddagger$
& \textbf{94.8} & 85.5 & \textbf{91.1}
& \textbf{94.8} & 58.5 & 76.6
& \textbf{93.5} & 57.3 & 75.4
& 81.0
& $\sim$20
& $\sim$22 \\

& ViCrop$^\ddagger$
& 90.4 & 85.5 & 88.5
& 91.0 & 59.8 & 75.4
& 88.3 & 56.8 & 72.5
& 78.8
& $\sim$20
& $\sim$22 \\

& Ours
& 90.4 & \textbf{89.5} & 90.1
& 90.8 & \textbf{67.3} & \textbf{79.0}
& 89.8 & \textbf{63.5} & \textbf{76.6}
& \textbf{81.9}
& \textbf{$\sim$5}
& \textbf{$\sim$16} \\

\noalign{\hrule height 0.8pt}
\end{tabular*}

\caption{
Comparison with HiDe and ViCrop under the same ZwZ-7B model.
Mean averages the three benchmark averages.
$^\ddagger$ denotes external evidence reacquisition using additional visual
inputs or visual re-encoding, whereas Base and Ours use a single visual
encoding pass.
Boldface indicates the better result between Base and Ours; tied values are
both highlighted.
Time denotes end-to-end wall-clock runtime on V$^*$Bench. For Ours, it
includes the unbatched and uncached text-only entity-extraction invocation and
the subsequent multimodal routing-and-answer invocation, both executed using
the same MLLM. Memory denotes dataset-averaged peak GPU usage; for two-GPU runs, it is the sum of the two GPU-wise average peak usages.
}
\label{tab:method_comparison}
\end{table*}

\subsection{Controlled Comparison of Compression-First Methods}
\label{app:controlled_pruning_comparison}

Table~\ref{tab:method_comparison} compares representative training-free methods on ZwZ-7B under the original high-resolution setting, providing complementary evidence on a different base model. We further conduct a controlled comparison on Qwen2.5-VL-7B with \texttt{max\_pixel}=1280 to reduce differences caused by the input budget and token-retention configuration. All generic pruning and merging baselines retain 33.3\% of the visual tokens. Under the same base model, visual-input budget, and retention ratio, every method completes without sample-level OOM. We consequently compare accuracy and end-to-end runtime under matched
conditions.

\begin{table}[t]
  \centering
  \scriptsize
  \setlength{\tabcolsep}{2.2pt}
  \renewcommand{\arraystretch}{1.12}

  \resizebox{\textwidth}{!}{%
  \begin{tabular}{@{}lcccccccccc@{}}
    \toprule
    \multirow{2}{*}{\textbf{Method}}
    & \multicolumn{3}{c}{\textbf{V$^*$Bench}}
    & \multicolumn{3}{c}{\textbf{HR-4K}}
    & \multicolumn{3}{c}{\textbf{HR-8K}}
    & \multirow{2}{*}{\textbf{Time}} \\
    \cmidrule(lr){2-4}
    \cmidrule(lr){5-7}
    \cmidrule(lr){8-10}
    & Attr & Spa. & Avg
    & FSP & FCP & Avg
    & FSP & FCP & Avg
    & \\
    \midrule

    \textbf{Base}
    & 72.2 & 65.8 & 69.6
    & 71.0 & 58.0 & 64.5
    & 57.3 & 53.0 & 55.1
    & \textbf{$\sim$3} \\

    \midrule

    HiPrune
    & \gaincell{73.0}{0.8}
    & \gaincell{71.1}{5.3}
    & \gaincell{72.3}{2.7}
    & \losscell{70.5}{0.5}
    & \gaincell{58.5}{0.5}
    & \samecell{64.5}
    & \losscell{55.3}{2.0}
    & \gaincell{\textbf{54.3}}{1.3}
    & \losscell{54.8}{0.3}
    & \textbf{$\sim$3} \\

    \addlinespace[1pt]

    VisionZip
    & \samecell{72.2}
    & \gaincell{71.1}{5.3}
    & \gaincell{71.7}{2.1}
    & \losscell{70.8}{0.2}
    & \losscell{57.5}{0.5}
    & \losscell{64.1}{0.4}
    & \losscell{56.5}{0.8}
    & \losscell{52.8}{0.2}
    & \losscell{54.6}{0.5}
    & $\sim$4 \\

    \addlinespace[1pt]

    V2Drop
    & \gaincell{73.0}{0.8}
    & \gaincell{75.0}{9.2}
    & \gaincell{73.8}{4.2}
    & \losscell{70.0}{1.0}
    & \losscell{55.0}{3.0}
    & \losscell{62.5}{2.0}
    & \losscell{56.8}{0.5}
    & \losscell{51.0}{2.0}
    & \losscell{53.9}{1.2}
    & \textbf{$\sim$3} \\

    \addlinespace[1pt]

    BTP
    & \losscell{64.3}{7.9}
    & \losscell{60.5}{5.3}
    & \losscell{62.8}{6.8}
    & \losscell{69.3}{1.7}
    & \losscell{51.3}{6.7}
    & \losscell{60.3}{4.2}
    & \losscell{56.8}{0.5}
    & \losscell{51.3}{1.7}
    & \losscell{54.0}{1.1}
    & $\sim$8 \\

    \addlinespace[1pt]

    TRIO
    & \losscell{70.4}{1.8}
    & \gaincell{69.7}{3.9}
    & \gaincell{70.2}{0.6}
    & \losscell{69.5}{1.5}
    & \gaincell{58.5}{0.5}
    & \losscell{64.0}{0.5}
    & \losscell{57.0}{0.3}
    & \samecell{53.0}
    & \losscell{55.0}{0.1}
    & \textbf{$\sim$3} \\

    \midrule

    \textbf{Ours}
    & \gaincell{\textbf{75.7}}{3.5}
    & \gaincell{\textbf{76.3}}{10.5}
    & \gaincell{\textbf{75.9}}{6.3}
    & \gaincell{\textbf{74.3}}{3.3}
    & \gaincell{\textbf{61.0}}{3.0}
    & \gaincell{\textbf{67.6}}{3.1}
    & \gaincell{\textbf{58.5}}{1.2}
    & \gaincell{\textbf{54.3}}{1.3}
    & \gaincell{\textbf{56.4}}{1.3}
    & \textbf{$\sim$3} \\

    \bottomrule
  \end{tabular}%
  }

    \caption{
    Controlled comparison on Qwen2.5-VL-7B with
    \texttt{max\_pixel}=1280. All compression-first pruning and merging
    baselines retain 33.3\% of the visual tokens. Green and red arrows denote
    absolute percentage-point improvements and decreases relative to Base,
    respectively; gray denotes no change. The reduced input budget allows all
    methods to complete without sample-level OOM. Time denotes end-to-end wall-clock
    runtime on V$^*$Bench in minutes. For Ours, it includes the unbatched and
    uncached text-only entity-extraction invocation and the subsequent multimodal
    routing-and-answer invocation, both executed using the same MLLM.
    }
  \label{tab:pruning_comparison_qwen25}
\end{table}

Under this controlled setting, Thinking-Once achieves the highest average accuracy on V$^*$Bench, HRBench-4K, and HRBench-8K, while remaining in the fastest runtime group at approximately three minutes. Generic token-reduction methods occasionally improve individual subsets, but their gains do not transfer consistently across the three high-resolution benchmarks. These results complement Table~\ref{tab:method_comparison} by showing that the advantage of Thinking-Once is not caused by competing methods failing under the original high-resolution setting or requiring a different device configuration.

\subsection{Additional Ablations}

Tables~\ref{tab:ablation_zwz7} and~\ref{tab:ablation_qwen3} report additional component ablations on ZwZ-7B and Qwen3-VL-8B, respectively. They complement the main ablation study by evaluating whether background context, denoising, entity-only routing, and global-only routing contribute consistently across different MLLMs.

\begin{table*}[t]
  \centering
  \small
  \setlength{\tabcolsep}{1.2pt}
  \renewcommand{\arraystretch}{1.08}

  \resizebox{\textwidth}{!}{%
  \begin{tabular}{@{}lccccccccccc@{}}
    \toprule
    \multirow{2}{*}{\textbf{Variant}}
    & \multicolumn{3}{c}{\textbf{V$^*$Bench}}
    & \multicolumn{3}{c}{\textbf{HR-4K}}
    & \multicolumn{3}{c}{\textbf{HR-8K}}
    & \multirow{2}{*}{\textbf{Mean}}
    & \multirow{2}{*}{\textbf{$\Delta$}} \\
    \cmidrule(lr){2-4}
    \cmidrule(lr){5-7}
    \cmidrule(lr){8-10}
    & Attr & Spa. & Avg
    & FSP & FCP & Avg
    & FSP & FCP & Avg
    & & \\
    \midrule

    Base
    & 90.4
    & 84.2
    & 88.0
    & 90.0
    & 59.5
    & 74.8
    & 88.5
    & 58.3
    & 73.4
    & 78.7
    & -3.2 \\

    \midrule

    \textbf{Full}
    & \textbf{90.4}
    & \textbf{89.5}
    & \textbf{90.1}
    & \textbf{90.8}
    & \textbf{67.3}
    & \textbf{79.0}
    & \textbf{89.8}
    & \textbf{63.5}
    & \textbf{76.6}
    & \textbf{81.9}
    & -- \\

    \addlinespace[1pt]

    w/o bg.
    & \samecell{90.4}
    & \losscell{86.8}{2.7}
    & \losscell{89.0}{1.1}
    & \losscell{89.8}{1.0}
    & \losscell{66.8}{0.5}
    & \losscell{78.3}{0.7}
    & \losscell{88.8}{1.0}
    & \losscell{63.3}{0.2}
    & \losscell{76.0}{0.6}
    & \losscell{81.1}{0.8}
    & \textcolor{lossRed}{$\downarrow$\,0.8} \\

    \addlinespace[1pt]

    Denoise
    & \samecell{90.4}
    & \losscell{85.5}{4.0}
    & \losscell{88.5}{1.6}
    & \losscell{90.0}{0.8}
    & \losscell{64.3}{3.0}
    & \losscell{77.1}{1.9}
    & \losscell{88.8}{1.0}
    & \losscell{62.0}{1.5}
    & \losscell{75.4}{1.2}
    & \losscell{80.3}{1.6}
    & \textcolor{lossRed}{$\downarrow$\,1.6} \\

    \addlinespace[1pt]

    Entity only
    & \samecell{90.4}
    & \losscell{82.9}{6.6}
    & \losscell{87.4}{2.7}
    & \losscell{89.5}{1.3}
    & \losscell{64.5}{2.8}
    & \losscell{77.0}{2.0}
    & \losscell{88.3}{1.5}
    & \losscell{62.8}{0.7}
    & \losscell{75.7}{0.9}
    & \losscell{80.0}{1.9}
    & \textcolor{lossRed}{$\downarrow$\,1.9} \\

    \addlinespace[1pt]

    Global only
    & \losscell{88.7}{1.7}
    & \losscell{85.5}{4.0}
    & \losscell{87.4}{2.7}
    & \losscell{90.0}{0.8}
    & \losscell{60.5}{6.8}
    & \losscell{75.3}{3.7}
    & \losscell{87.8}{2.0}
    & \losscell{58.3}{5.2}
    & \losscell{73.0}{3.6}
    & \losscell{78.6}{3.3}
    & \textcolor{lossRed}{$\downarrow$\,3.3} \\

    \bottomrule
  \end{tabular}%
  }

  \caption{
  Component ablations on ZwZ-7B. Green and red arrows denote absolute
  percentage-point improvements and decreases relative to Full, respectively;
  gray denotes no change. Mean averages the three benchmark averages, and
  $\Delta$ reports the corresponding Mean change relative to Full.
  }
  \label{tab:ablation_zwz7}
\end{table*}

\begin{table*}[t]
  \centering
  \small
  \setlength{\tabcolsep}{1.2pt}
  \renewcommand{\arraystretch}{1.08}

  \resizebox{\textwidth}{!}{%
  \begin{tabular}{@{}lccccccccccc@{}}
    \toprule
    \multirow{2}{*}{\textbf{Variant}}
    & \multicolumn{3}{c}{\textbf{V$^*$Bench}}
    & \multicolumn{3}{c}{\textbf{HR-4K}}
    & \multicolumn{3}{c}{\textbf{HR-8K}}
    & \multirow{2}{*}{\textbf{Mean}}
    & \multirow{2}{*}{\textbf{$\Delta$}} \\
    \cmidrule(lr){2-4}
    \cmidrule(lr){5-7}
    \cmidrule(lr){8-10}
    & Attr & Spa. & Avg
    & FSP & FCP & Avg
    & FSP & FCP & Avg
    & & \\
    \midrule

    Base
    & 84.4
    & 80.3
    & 82.7
    & 91.3
    & 65.8
    & 78.5
    & 84.0
    & 62.8
    & 73.4
    & 78.2
    & -2.0 \\

    \midrule

    \textbf{Full}
    & \textbf{87.8}
    & \textbf{81.6}
    & \textbf{85.3}
    & \textbf{91.8}
    & \textbf{70.5}
    & \textbf{81.1}
    & \textbf{84.3}
    & \textbf{64.3}
    & \textbf{74.3}
    & \textbf{80.2}
    & -- \\

    \addlinespace[1pt]

    w/o bg.
    & \losscell{84.4}{3.4}
    & \losscell{80.3}{1.3}
    & \losscell{82.7}{2.6}
    & \losscell{91.5}{0.3}
    & \losscell{67.3}{3.2}
    & \losscell{79.4}{1.7}
    & \losscell{83.8}{0.5}
    & \losscell{63.3}{1.0}
    & \losscell{73.5}{0.8}
    & \losscell{78.5}{1.7}
    & \textcolor{lossRed}{$\downarrow$\,1.7} \\

    \addlinespace[1pt]

    Denoise
    & \losscell{84.4}{3.4}
    & \losscell{80.3}{1.3}
    & \losscell{82.7}{2.6}
    & \losscell{91.3}{0.5}
    & \losscell{66.0}{4.5}
    & \losscell{78.6}{2.5}
    & \losscell{84.0}{0.3}
    & \losscell{62.8}{1.5}
    & \losscell{73.4}{0.9}
    & \losscell{78.2}{2.0}
    & \textcolor{lossRed}{$\downarrow$\,2.0} \\

    \addlinespace[1pt]

    Entity only
    & \losscell{87.0}{0.8}
    & \losscell{80.3}{1.3}
    & \losscell{84.3}{1.0}
    & \losscell{91.3}{0.5}
    & \losscell{66.3}{4.2}
    & \losscell{78.8}{2.3}
    & \samecell{84.3}
    & \losscell{63.8}{0.5}
    & \losscell{74.0}{0.3}
    & \losscell{79.0}{1.2}
    & \textcolor{lossRed}{$\downarrow$\,1.2} \\

    \addlinespace[1pt]

    Global only
    & \losscell{82.6}{5.2}
    & \losscell{73.7}{7.9}
    & \losscell{79.1}{6.2}
    & \losscell{91.3}{0.5}
    & \losscell{65.0}{5.5}
    & \losscell{78.1}{3.0}
    & \losscell{81.8}{2.5}
    & \losscell{63.3}{1.0}
    & \losscell{72.5}{1.8}
    & \losscell{76.6}{3.7}
    & \textcolor{lossRed}{$\downarrow$\,3.7} \\

    \bottomrule
  \end{tabular}%
  }

  \caption{
  Component ablations on Qwen3-VL-8B. Green and red arrows denote absolute
  percentage-point improvements and decreases relative to Full, respectively;
  gray denotes no change. Mean averages the three benchmark averages, and
  $\Delta$ reports the corresponding Mean change relative to Full.
  }
  \label{tab:ablation_qwen3}
\end{table*}

\subsection{Qualitative Examples}

This subsection provides qualitative examples showing how Thinking-Once routes question-relevant evidence in high-resolution scenes. The enlarged regions and heatmaps in the figures are used only for visualization; the method itself does not introduce crops, re-encoded sub-images, or a second visual input during inference.

\begin{figure}[!p]
    \centering
    \includegraphics[height=0.86\textheight,keepaspectratio]{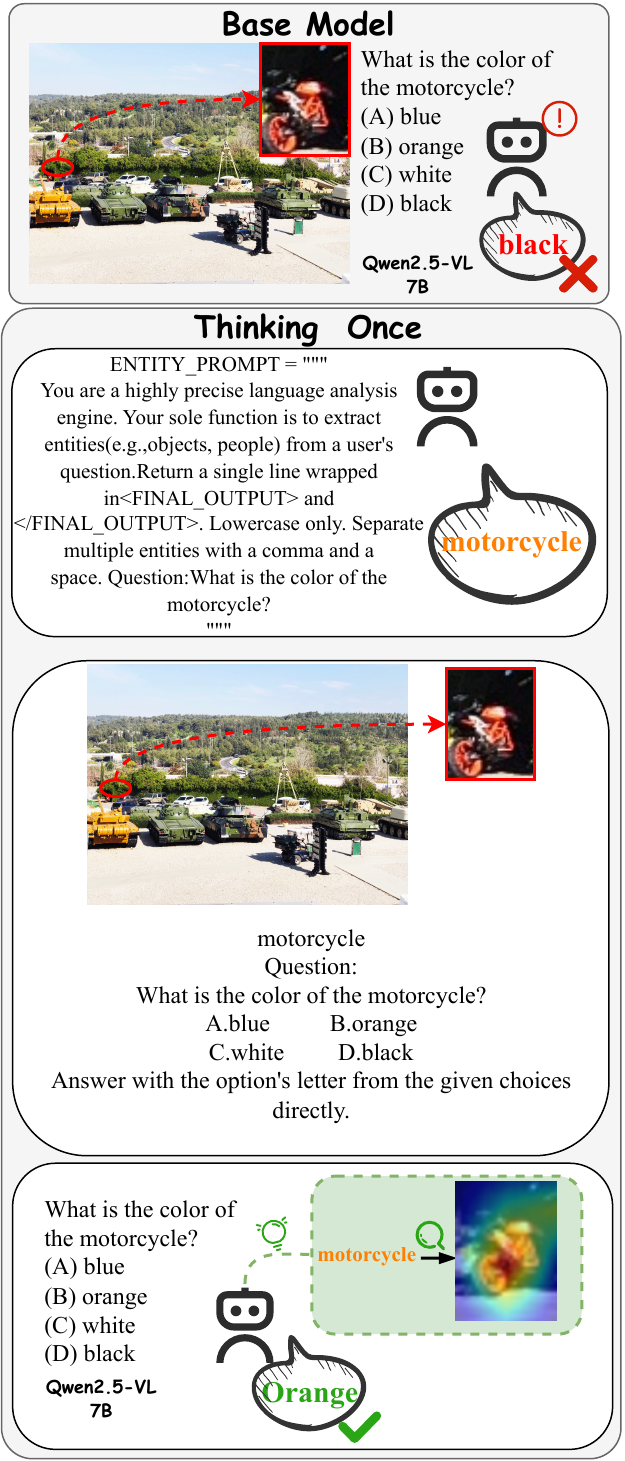}
    \caption{
    Single-object attribute case. The base model predicts the wrong color for a
    small motorcycle in a high-resolution scene, while Thinking-Once extracts the
    entity query, localizes the relevant evidence internally, and routes it to
    later layers for the correct answer. The zoomed patch and heatmap are
    visualization aids rather than additional visual inputs.
    }
    \label{fig:app_case_single_object}
\end{figure}

Figure~\ref{fig:app_case_single_object} shows a single-object attribute case. The base model fails to identify the color of a small motorcycle in a crowded high-resolution image. Thinking-Once first extracts the queried entity ``motorcycle'' from the question, uses it as an internal routing query, and localizes the corresponding evidence at intermediate layers. By routing this entity-centered evidence to later layers, the model answers the attribute question correctly.

\begin{figure}[!p]
    \centering
    \includegraphics[height=0.86\textheight,keepaspectratio]{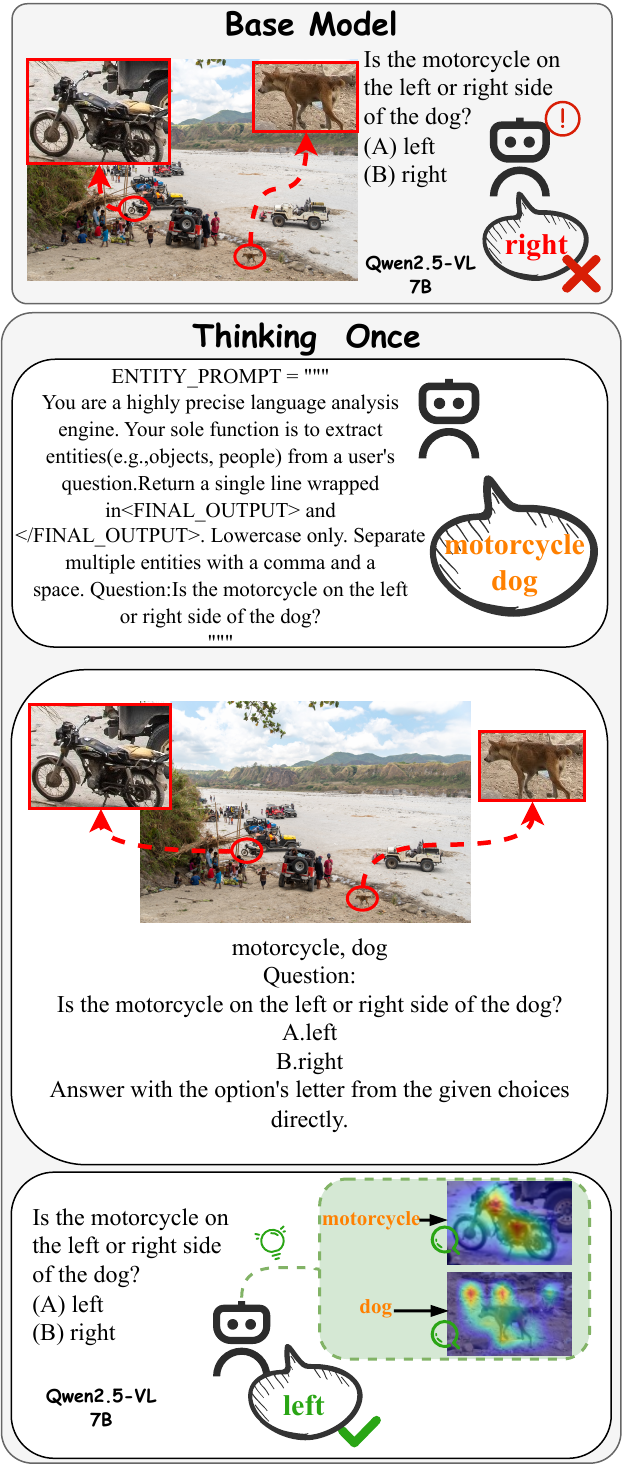}
    \caption{
    Multi-object relational case. The base model gives an incorrect spatial
    answer, whereas Thinking-Once extracts both queried entities, routes their
    question-relevant evidence independently, and preserves the evidence needed
    for relative-position reasoning. The visualized patches and heatmaps are for
    explanation only and are not used as extra inputs.
    }
    \label{fig:app_case_multi_object}
\end{figure}

Figure~\ref{fig:app_case_multi_object} shows a multi-object relational case. The question requires comparing the positions of two entities, namely the motorcycle and the dog. Thinking-Once extracts both entities and performs query-wise evidence routing so that each object can retain its own core evidence. This supports cross-instance spatial reasoning and leads to the correct relative-position answer.

The following qualitative examples further visualize layer-wise evidence localization from Layer 1 to Layer 28. They illustrate how question-guided responses evolve across depth and become more concentrated on the queried visual evidence in intermediate layers, supporting the evidence-routing window discussed in the main paper.

\begin{figure}[!p]
    \centering
    \includegraphics[width=0.96\textwidth]{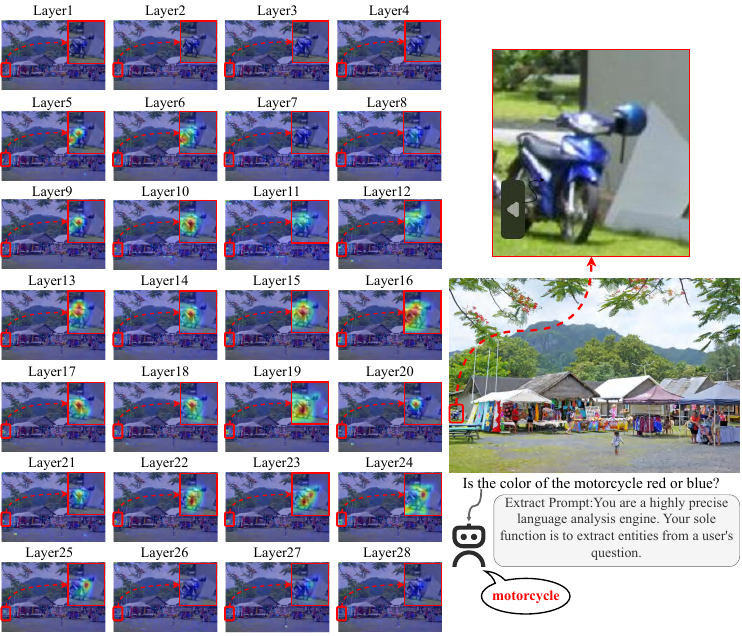}
    \caption{
    Layer-wise qualitative example of question-guided evidence localization.
    The visualization shows how the model response to the queried entity evolves
    across layers and becomes concentrated around the relevant visual evidence.
    }
    \label{fig:app_layerwise_evidence_1}
\end{figure}

\begin{figure}[!p]
    \centering
    \includegraphics[width=0.96\textwidth]{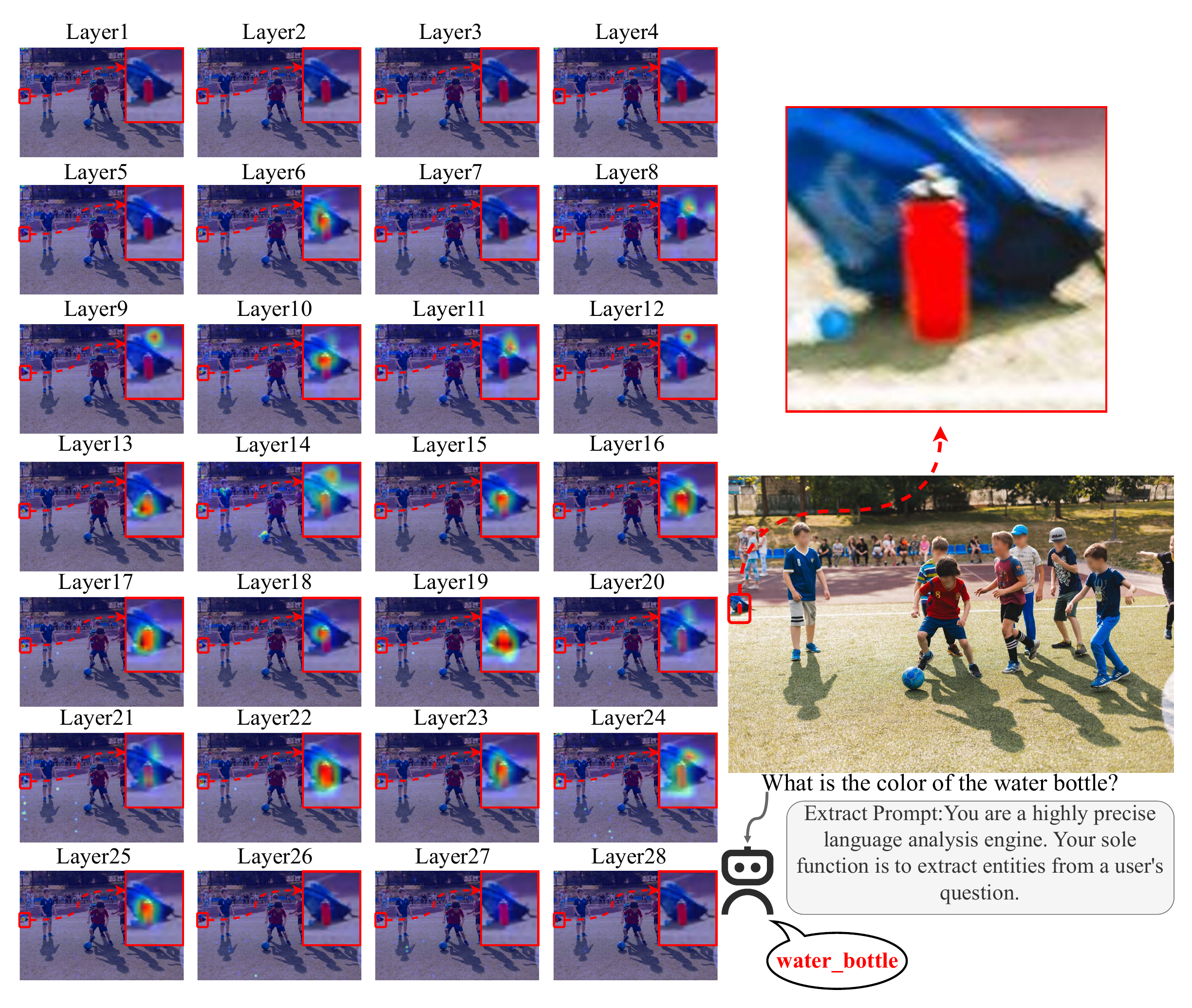}
    \caption{
    Additional layer-wise qualitative example of question-guided evidence
    localization. The example further illustrates that intermediate layers can
    expose question-relevant visual evidence before final answer generation.
    }
    \label{fig:app_layerwise_evidence_2}
\end{figure}

These qualitative examples show that question-relevant regions are not uniformly localized across all layers. Early layers preserve visual details but may lack stable question-conditioned focus, while late layers can absorb the evidence into downstream hidden states. The clearest localization typically appears in intermediate layers, which is consistent with the evidence-routing window identified by the oracle intervention analysis in the main paper.

\section{Extended Analysis}
\label{app:extended_analysis}

This appendix expands the evidence-allocation analysis behind Thinking-Once. The main paper presents the core empirical observations: visual information can survive encoding, entity-centered tokens have high causal value, useful routing is layer-dependent, and non-core tokens may still provide necessary context. Here we first define the cross-layer evidence-retention diagnostic used in the main paper, introduce a layer-wise object-attention enrichment analysis that directly measures whether target-centered attention survives deeper processing, and provide a matched-budget analysis of the two central routing choices. We then connect the broader observations more explicitly and state the boundaries under which single-visual-pass evidence routing is expected to help.

\subsection{Cross-Layer GT-Evidence Retention}
\label{app:cross_layer_retention}

We quantify how much downstream attention remains allocated to the same routed
GT evidence after the routing layer. For sample $s$, let $G_s$ be the set of
original visual tokens associated with its annotated GT region, let
$S_{\mathrm{core},s}$ be the core selected by Thinking-Once, and define the
fixed target set
\begin{equation}
T_s = G_s \cap S_{\mathrm{core},s}.
\label{app:eq:retention_target}
\end{equation}
The same target-token identities are used for both Base and Thinking-Once. For
Thinking-Once, their indices are remapped after sequence routing while their
original spatial identities are retained. Background-summary tokens are
included in the routed visual sequence but are not counted as members of
$T_s$.

At layer $\ell\geq L$, we reconstruct the complete causal attention row of the
global-question routing query. Let $\mathcal{K}^{\ell}_{c,s}$ be the key positions present
in condition $c\in\{\mathrm{Base},\mathrm{TO}\}$ at that layer and let
$z^{\ell}_{c,s}(i)$ be the corresponding attention logit. The layer-wise
retention mass is
\begin{equation}
M^{\ell}_{c,s}
=
\sum_{i\in T^{\ell}_{c,s}}
\frac{\exp z^{\ell}_{c,s}(i)}
{\sum_{k\in\mathcal{K}^{\ell}_{c,s}}\exp z^{\ell}_{c,s}(k)},
\label{app:eq:cross_layer_retention}
\end{equation}
where $T^{\ell}_{c,s}$ denotes the layer-specific indices of the fixed target
tokens. Thus, the metric tracks the share of the query's complete causal
attention assigned to the same full-resolution GT-core evidence, rather than
changing the target region independently for the two conditions.

For the cross-model visualization, we first average $M^{\ell}_{c,s}$ over
valid samples within each MLLM. Both trajectories of a model are then
divided by that model's maximum Base mass, and post-routing depth is mapped
to $d=(\ell-L)/(D-L)$. We linearly interpolate the trajectories to a common
depth grid before averaging across Qwen2.5-VL-7B, ZwZ-7B, and Qwen3-VL-8B. The
plotted three-point weighted moving average is used only for readability. To
test whether retention is associated with task success, we compute each
sample's mean downstream Thinking-Once-minus-Base mass, standardize it within
model, and compare correct ($n=505$) and incorrect ($n=68$) answers using a
two-sided Mann--Whitney test over all 573 model--sample pairs
($U=23{,}793$, $p\approx2.4\times10^{-7}$, rank-biserial $r=0.3857$).

Because Thinking-Once changes the set of keys after routing,
Eq.~\eqref{app:eq:cross_layer_retention} measures relative attention allocation
within each condition; it should not by itself be interpreted as a
token-count-invariant causal effect. We therefore use it as a mechanism
diagnostic and rely on the matched-budget controls below to separate evidence
identity and context organization from the number of retained tokens.

\subsection{Layer-Wise Object-Attention Enrichment}
\label{app:object_attention_enrichment}

The fixed-token retention metric above tracks attention assigned to the same full-resolution GT-core tokens after routing. We additionally introduce a spatial enrichment diagnostic that asks a complementary question: at each language-model layer, how concentrated is the model's visual attention inside the annotated target region relative to the fraction of the visual grid occupied by that region? This diagnostic is computed on all 191 V$^*$Bench examples with ground-truth object bounding boxes. Suppressing the sample index for readability, the enrichment at language-model layer $l$ is
\begin{equation}
E_l
=
\frac{
\sum_{i\in B} a_{l,i}\big/\sum_{i\in V}a_{l,i}
}{
|B|/|V|
},
\label{app:eq:object_attention_enrichment}
\end{equation}
where $V$ is the set of original visual tokens, $B\subseteq V$ contains the visual tokens overlapping the ground-truth bounding box, and $a_{l,i}$ denotes the fused routing-query-to-visual attention assigned to token $i$. Thus, $E_l=1$ corresponds to spatially uniform visual attention, whereas $E_l>1$ indicates preferential concentration inside the target region. For example, $E_l=100$ means that the attention density inside the target box is $100$ times the density expected under a uniform spatial allocation; it does not mean that the box receives $100\%$ of the total visual attention. We compute enrichment separately for each example and report the sample mean, with shaded regions showing $\pm1$ standard error of the mean. Because the values span several orders of magnitude, the vertical axis is logarithmic.

\begin{figure}[!htbp]
    \centering
    \includegraphics[width=0.98\textwidth]{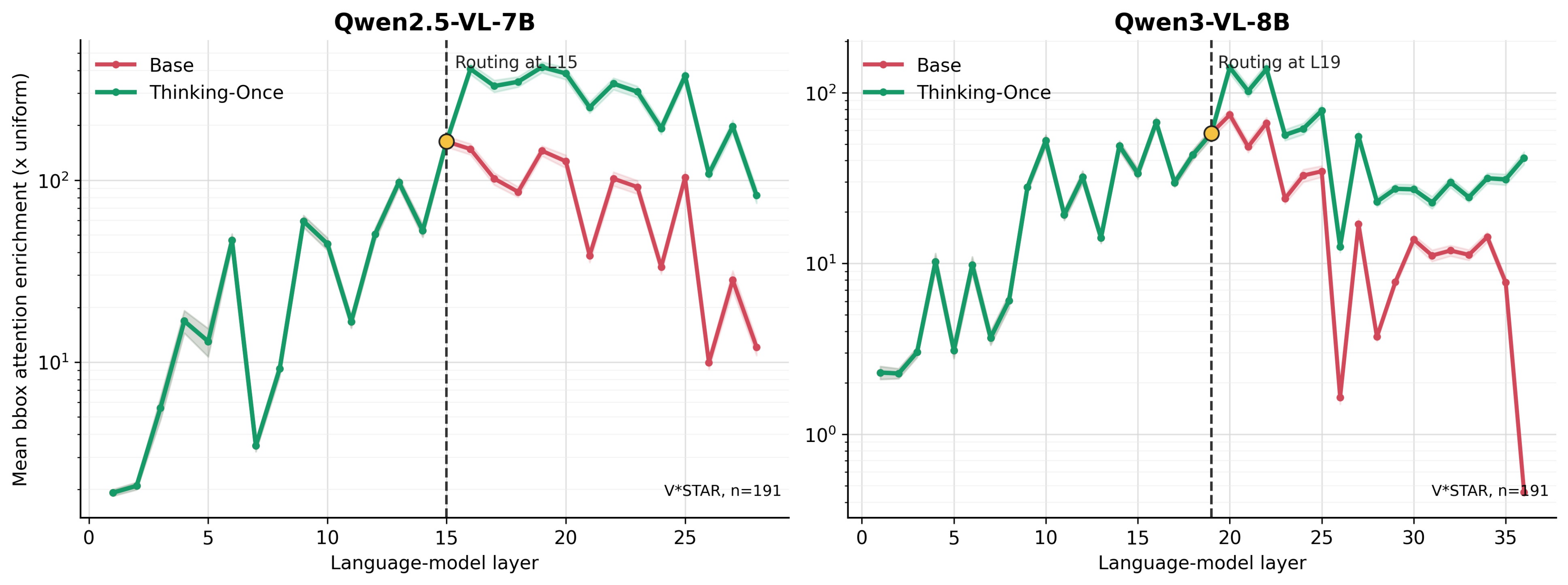}
    \caption{
    Layer-wise object-attention enrichment on V$^*$Bench. The two panels compare Base and Thinking-Once on all 191 examples with ground-truth object boxes. Enrichment normalizes the fraction of visual attention inside the target box by the fraction of visual tokens covered by that box, so $1\times$ denotes uniform spatial attention and values above $1\times$ denote target-centered concentration. Curves show sample means, shaded regions indicate $\pm1$ standard error, and the vertical axis is logarithmic. Base and Thinking-Once use identical inputs and routing queries, causing exact pre-routing overlap. Dashed lines and yellow markers indicate routing after L15 for Qwen2.5-VL-7B and after L19 for Qwen3-VL-8B; the first affected layers are therefore L16 and L20. Post-routing attention is projected from the compact core-plus-background representation back to the original visual grid in a mass-preserving manner, with each background summary redistributed only over its original source set $U_g$. Ground-truth boxes are used only for evaluation, not for routing.
    }
    \label{fig:app_object_attention_enrichment}
\end{figure}

\paragraph{Controlled routing comparison.}
The Base and Thinking-Once runs use exactly the same image, cleaned question, routing prompt, and routing-query tokens. The only difference is that sequence routing is disabled for the Base control. Their trajectories therefore overlap exactly before the routing point; in Figure~\ref{fig:app_object_attention_enrichment}, the red Base curve is visually occluded by the green Thinking-Once curve in these layers. The dashed lines and yellow markers denote the outputs of the routing layers: L15 for Qwen2.5-VL-7B and L19 for Qwen3-VL-8B. Consistent with the layer convention used throughout this appendix, routing is applied \emph{after} the marked layer, so the first affected layers are L16 and L20, respectively.

For this diagnostic, Thinking-Once uses independent top-$p$ routing with $\rho=0.7$ to select full-resolution core evidence and retains global scene context through background summaries constructed over an $8\times8$ spatial grid. The compact representation is propagated through the remaining language-model layers without cropping, image re-encoding, or a second visual forward pass. To evaluate post-routing spatial attention on the original grid, attention over the compact representation is projected back in a mass-preserving manner: each retained core token is assigned to its original location, and each background-summary token is redistributed only over its original source set $U_g$, namely the unselected visual tokens summarized within the corresponding spatial cell, while preserving its total attention mass. Ground-truth boxes are used only to compute Eq.~\eqref{app:eq:object_attention_enrichment}; they are never used to select routed tokens.

\FloatBarrier

\paragraph{Observed retention.}
For Qwen2.5-VL-7B, both conditions reach an enrichment of $163.0\times$ at L15. Immediately after routing, Thinking-Once increases the enrichment to $408.3\times$ at L16, compared with $148.1\times$ for Base. The Base model then loses much of its target-centered attention and ends at $12.1\times$ at L28, whereas Thinking-Once retains $82.5\times$, corresponding to a $6.8\times$ final-layer improvement. Qwen3-VL-8B exhibits the same pattern. Both conditions reach $57.9\times$ at L19; at L20, Thinking-Once reaches $140.4\times$, compared with $74.1\times$ for Base. At the final layer, Base falls to $0.46\times$, below the uniform-attention reference, while Thinking-Once maintains $41.5\times$, a $90.5\times$ relative improvement.

\paragraph{Interpretation.}
Figure~\ref{fig:app_object_attention_enrichment} provides mechanistic evidence for the motivation behind Thinking-Once. In both model families, substantial bounding-box enrichment has already emerged at intermediate layers, showing that the relevant visual evidence is not necessarily absent or imperceptible. The limitation of the Base model is instead that this object-centered allocation deteriorates and fluctuates strongly during deeper processing. Thinking-Once intervenes precisely when object-specific evidence is accessible, converts the visual sequence into high-confidence core tokens plus compact global context, and routes this representation through the remaining layers. The persistent post-routing separation between the green and red curves indicates that compact evidence routing improves the retention of target-centered visual information across architectures.

Importantly, the curves diverge only after routing. This controlled temporal separation rules out differences in the input image, prompt, query construction, or pre-routing hidden states as explanations for the effect. The figure should therefore be interpreted as evidence for \emph{attention retention and evidence utilization}, rather than as a direct accuracy measurement. Together with the answer-accuracy results and the fixed-token retention diagnostic above, it supports the central claim that Thinking-Once prevents already encoded visual evidence from being diluted or displaced before answer generation.

\subsection{Matched-Budget Evidence Coverage Analysis}
\label{app:matched_budget_coverage}

The main method combines two allocation decisions: how to select the full-resolution core evidence and how to preserve compact contextual support around that core. Because either decision can change the number of routed tokens, an uncontrolled comparison could attribute improved evidence coverage to a larger token budget rather than to a better routing rule. We therefore construct two matched-budget experiments that isolate the effects of query selection and context organization while keeping the corresponding token counts fixed.

\paragraph{Metrics.}
For each sample, let $G_s$ denote the visual tokens associated with the annotated ground-truth evidence region, and let $S_s$ denote the selected full-resolution core tokens. We measure GT-evidence recall as
\begin{equation}
\mathrm{Recall}_{\mathrm{GT}}
=
\frac{1}{|\mathcal{D}|}
\sum_{s\in\mathcal{D}}
\frac{|S_s\cap G_s|}{|G_s|}.
\label{app:eq:gt_evidence_recall}
\end{equation}
To measure whether the retained context remains spatially distributed, we partition the original visual-token map into the same regular $8\times8$ grid used by Thinking-Once. Spatial-context coverage is the average percentage of grid cells represented by at least one retained or summarized context element. This metric evaluates spatial support rather than task accuracy, and is used only to diagnose how different allocations distribute the same context budget.

\paragraph{Matched core-token budget.}
For every sample $s$, Independent first performs minimum-coverage selection separately for each routing query and forms the union of the resulting sets. Its resulting size $K_s$ is then used as the exact per-sample core-token budget for all query-selection variants. Thus, every variant returns exactly $K_s$ core tokens, and differences in GT-evidence recall cannot be explained by retaining more tokens. The variants are defined as follows:
\begin{itemize}
    \item \textbf{Random} uniformly samples $K_s$ visual tokens without using the question.
    \item \textbf{Global} ranks visual tokens using the global question query and retains its top-$K_s$ tokens.
    \item \textbf{Average} averages the normalized evidence distributions of all routing queries and retains the top-$K_s$ tokens under the aggregated distribution.
    \item \textbf{Max} takes the element-wise maximum across the normalized routing-query distributions and retains the top-$K_s$ tokens.
    \item \textbf{Independent} performs minimum-coverage selection for each routing query separately and takes the union, as used by Thinking-Once.
\end{itemize}

\paragraph{Matched total-token budget.}
For the context experiment, the core set is held fixed across all variants, with an average of $233.42$ core tokens per sample. Each sample uses the same matched context budget across variants, which differ only in how that context budget is allocated:
\begin{itemize}
    \item \textbf{Extra Core} assigns the additional budget to the highest-ranked non-core tokens, without explicitly preserving spatial distribution.
    \item \textbf{Random Context} samples non-core context tokens uniformly at random.
    \item \textbf{Uniform Context} selects non-core tokens at approximately uniform spatial intervals over the visual-token map.
    \item \textbf{Grid Background} partitions the visual-token map into an $8\times8$ grid and summarizes the unselected tokens within each spatial cell, as used by Thinking-Once.
\end{itemize}

\begin{table}[!htbp]
\centering
\small
\setlength{\tabcolsep}{8.0pt}
\renewcommand{\arraystretch}{1.08}

\begin{tabular}{lcc}
\toprule
\textbf{Group} & \textbf{Variant} & \textbf{Coverage (\%)} \\
\midrule
\multirow{5}{*}{\makecell[l]{Query routing\\GT-evidence recall}}
& Random      & 4.87 \\
& Global      & 4.66 \\
& Average     & 71.13 \\
& Max         & 69.22 \\
& \textbf{Independent} & \textbf{85.26} \\
\midrule
\multirow{4}{*}{\makecell[l]{Context allocation\\Spatial coverage}}
& Extra Core      & 80.61 \\
& Random Context  & 89.58 \\
& Uniform Context & 92.37 \\
& \textbf{Grid Background} & \textbf{100.00} \\
\bottomrule
\end{tabular}

\caption{
Matched-budget evidence-coverage analysis. Query-routing variants use the same per-sample core-token count, while context-allocation variants use the same fixed core set and total routed-token budget. Higher values indicate more complete coverage of the corresponding evidence structure.
}
\label{tab:app_matched_budget_coverage}
\end{table}

Table~\ref{tab:app_matched_budget_coverage} shows that independent routing provides substantially more complete access to the annotated evidence than query-agnostic or query-aggregated alternatives. Random and Global retain only $4.87\%$ and $4.66\%$ of the GT evidence, respectively, indicating that a single generic selection signal is insufficient for questions with multiple semantic demands. Average and Max recover much more evidence, but their aggregation can still suppress tokens that are important to only one entity or one question aspect. Independent reaches $85.26\%$ recall, improving over the stronger aggregated baseline Average by $14.13$ percentage points. This result supports the per-query coverage constraint in Eq.~\eqref{app:eq:canonical_set}: each routing query should preserve its own evidence before the selected sets are merged.

The context comparison exhibits a complementary pattern. Extra Core covers $80.61\%$ of the spatial cells, showing that assigning more tokens to already salient regions does not guarantee broad contextual support. Random Context and Uniform Context increase coverage to $89.58\%$ and $92.37\%$, respectively, but can still leave parts of the visual field unsupported. Grid Background reaches $100.00\%$ coverage under the same total budget, exceeding Uniform Context by $7.63$ percentage points. The gain therefore comes from organizing context according to the image's spatial structure, rather than from increasing the number of routed tokens.

Together, the two controlled comparisons isolate the roles of the method's two main components. Independent minimum-coverage routing protects evidence associated with distinct semantic queries, while structured grid summaries preserve compact support across the full visual field. Their advantages under matched budgets show that Thinking-Once benefits from better evidence allocation, not merely from retaining more tokens.

\subsection{Complementary Evidence that Useful Information Already Survives Encoding}
\label{app:complementary_evidence}

The main text uses LongCat-Next
\cite{meituanlongcatteam2026longcatnextlexicalizingmodalitiesdiscrete}
to show that lightweight decoders can reconstruct image structure, object boundaries, and spatial layout from frozen vision-encoder outputs. This does not prove that every high-resolution question is answerable from the encoded representation. Instead, it establishes a narrower but necessary condition for single-visual-pass routing: local visual information can survive visual encoding and remain available to later multimodal layers.

Evidence from interleaved visual reasoning and vision tool-use training supports the same distinction from another direction. Yang et al.~\cite{yang2026position} show that removing interleaved images from several visual CoT systems often causes only small performance changes, suggesting that intermediate images are not always the dominant source of the observed gain. DeepEyes further shows that text-only chain-of-thought training can already improve high-resolution VQA over the base model~\cite{zheng2025deepeyes}, indicating that part of the improvement may come from reasoning alignment or task adaptation rather than newly acquired visual observations alone. Similarly, Ma et al.~\cite{ma2026does} decompose crop-and-zoom tool-use reinforcement learning and find that a large part of the training progress can be attributed to tool-free intrinsic improvement.

These findings do not imply that crops, zoomed views, or visual tools are unnecessary. They instead show that performance gains from such systems should not be attributed solely to additional visual evidence. In many cases, improved performance may also come from better use of information already present in the original visual encoding or in intermediate hidden states. This is consistent with recent evidence that large models may internally encode relevant evidence even when it is not effectively surfaced in the final prediction
\cite{liu2025selfelicit}. Thinking-Once is designed for this
utilization-limited regime: the required evidence is at least partially encoded, but it is diluted, misallocated, or discarded before answer generation.

\subsection{From Oracle Interventions to an Evidence-Routing Window}
\label{app:routing_window_extended}

The layer-wise oracle interventions in the main paper separate three factors that are otherwise entangled: token budget, token identity, and intervention depth. \emph{Random Tokens} preserve the same number of tokens as the entity-region tokens, \emph{Oracle Core Tokens} preserve only tokens inside the ground-truth box, and \emph{Remove Oracle Tokens} preserves all tokens except those inside the box. This design tests whether performance depends merely on using fewer tokens, or on preserving the causal identity of the evidence.

The comparisons provide two complementary signals. The advantage of Oracle Core Tokens over Random Tokens shows a sufficiency signal: a compact entity-centered subset can support strong reasoning when it contains the relevant evidence. The failure of Remove Oracle Tokens gives a necessity signal: abundant non-target information cannot reliably replace evidence removed from the entity region. Because Random Tokens use the same budget as Oracle Core Tokens, and Remove Oracle Tokens keeps substantially more tokens, the result cannot be explained by sequence length alone. What matters is which tokens are routed and when their influence is redirected.

The layer dependence of these interventions further defines an
\emph{evidence-routing window}. Early layers may preserve fine visual detail but
lack stable question-conditioned correspondence. Late layers contain richer multimodal states, but too little computation remains for a routing operation to change the reasoning trajectory. Intermediate layers provide the useful overlap: question-relevant evidence has become identifiable, while enough downstream computation remains for the routed evidence to affect the answer. The earliest oracle optimum should therefore not be interpreted as the practical routing layer, because ground-truth boxes provide perfect localization that is unavailable at inference time. A usable routing layer must jointly provide reliable question-conditioned localization and sufficient remaining depth.

\subsection{Why the Evidence Set Must Preserve Context}
\label{app:structured_context_extended}

Entity tokens form the evidence core, but the core alone is not always sufficient. The background-masking analysis in the main paper shows a non-monotonic effect: moderate background removal can improve accuracy, while aggressive masking degrades it. This indicates that background tokens contain a mixture of interference and support. Some background tokens are redundant or harmful, but others encode spatial references, relational cues, or attribute-comparison evidence needed to interpret the target entity.

This observation explains why object-only cropping can be brittle in HR-VQA. A question may refer to a single entity while asking about its relation to another object, its relative position, or an attribute comparison whose evidence lies outside the entity box. Removing all surrounding tokens preserves the referent but can destroy part of the structure required to answer the question. Thus, the routing objective should not be to keep the smallest possible entity-only crop. It should preserve the smallest evidence structure that still supports the question.

Attention-sink interventions reveal a related form of context. Although attention-sink tokens can distort attention-based localization, suppressing them reduces accuracy across models and benchmarks in the main analysis. This shows that a token that is spatially diffuse or visually non-discriminative can still carry useful hidden-state information, such as aggregation or information transfer signals. Together, background masking and sink suppression indicate that effective HR-VQA evidence is structured rather than merely sparse: it consists of a dense entity-centered core plus a compact support system that preserves explicit and implicit context.

\subsection{Applicability Boundaries and Falsifiability}
\label{app:falsifiability}

The evidence-allocation account is intended for a utilization-limited regime, not for all HR-VQA failures. Routing cannot recover details that the vision encoder has irreversibly removed, nor can it correct an initial encoding that never represents the relevant region. It may also provide limited benefit when the base model already allocates evidence reliably, or when the question requires new visual observations outside the original input. In such cases, cropping, zooming, or multi-round visual search remains appropriate. Thinking-Once is therefore complementary to evidence reacquisition methods: it addresses failures caused by inefficient preservation and allocation of available evidence, rather than failures caused by absent evidence.

This account makes several testable predictions. First, routing should help more on questions involving relations, comparisons, or cross-instance context than on questions solvable from an isolated attribute, because such questions require both a precise entity anchor and supporting context. Second, gains should be larger for crowded high-resolution inputs, where relevant evidence competes with many irrelevant tokens and is more vulnerable to dilution. Third, interventions should be most effective at intermediate layers, after question-conditioned correspondence has emerged but before downstream reasoning has become too close to the output. Fourth, removing all context or suppressing sink states should hurt performance when those tokens carry useful support signals.

The account would be weakened if these patterns consistently failed to appear. For example, if routing mainly improved isolated attribute questions, if gains were unrelated to visual clutter, if query-agnostic pruning at arbitrary layers matched the routing gains, or if context removal never hurt performance, then the proposed evidence-allocation explanation would be insufficient. These criteria make the analysis not only a motivation for Thinking-Once, but also a falsifiable description of when and why single-visual-pass routing should work.

\bibliographystyle{aaai2027}
\bibliography{aaai2027}